\documentclass[runningheads]{llncs}

\usepackage[year=2024]{eccv}
\usepackage{eccvabbrv}

\usepackage{graphicx}
\usepackage{booktabs}

\usepackage[accsupp]{axessibility}  % Improves PDF readability for those with disabilities.

\usepackage{multirow}
\usepackage[pagebackref,breaklinks,colorlinks,citecolor=eccvblue]{hyperref}
\usepackage{orcidlink}
\newlength\savewidth\newcommand\shline{\noalign{\global\savewidth\arrayrulewidth
  \global\arrayrulewidth 1pt}\hline\noalign{\global\arrayrulewidth\savewidth}}

\renewcommand{\paragraph}[1]{\vspace{1.25mm}\noindent\textbf{#1}}

\begin{document}
\title{How Powerful Potential of Attention on \\ Image Restoration? 
% Continuous Scaling Attention Can Beat CNNs and Transformers on Image Restoration
} 

% TODO REVIEW: If the paper title is too long for the running head, you can set How Powerful Potential of Attention on Image Restoration
\titlerunning{How Powerful Potential of Attention on 
\\
Image Restoration? }

\author{Cong Wang\inst{1,2} \and
Jinshan Pan\inst{3} \and
Yeying Jin\inst{4} \and
Liyan Wang\inst{5} \and
Wei Wang\inst{1} \and
Gang Fu\inst{2} \and
Wenqi Ren\inst{1} \and
Xiaochun Cao\inst{1}}

\authorrunning{Cong~Wang et al.}

\institute{
Sun Yat-sen University \and
The Hong Kong Polytechnic University \and
Nanjing University of Science and Technology \and
National University of Singapore \and
Dalian University of Technology
}

\maketitle

\begin{abstract}
Transformers have demonstrated their effectiveness in image restoration tasks. Existing Transformer architectures typically comprise two essential components: multi-head self-attention and feed-forward network (FFN). The former captures long-range pixel dependencies, while the latter enables the model to learn complex patterns and relationships in the data. Previous studies have demonstrated that FFNs are key-value memories~\cite{geva2020transformer}, which are vital in modern Transformer architectures. In this paper, we conduct an empirical study to explore the potential of attention mechanisms without using FFN and provide novel structures to demonstrate that removing FFN is flexible for image restoration. Specifically, we propose Continuous Scaling Attention (\textbf{CSAttn}), a method that computes attention continuously in three stages without using FFN. To achieve competitive performance, we propose a series of key components within the attention. Our designs provide a closer look at the attention mechanism and reveal that some simple operations can significantly affect the model performance. We apply our \textbf{CSAttn} to several image restoration tasks and show that our model can outperform CNN-based and Transformer-based image restoration approaches. 
% The source codes will be made available.
 
\end{abstract}

\vspace{-3mm}
\section{Introduction}
\vspace{-1mm}

Image restoration aims to recover clear images from the given degraded ones.
Reconstructing high-quality images benefits many practical applications, such as video surveillance, and self-driving techniques. 
This task is highly ill-posed as only degraded images are known while many solutions can correspond to the original ones. 
Early studies always apply statistic priors to make the problems well-posed~\cite{he2010single,pan_pami_l0,Huang_2015_CVPR,pan2016blind}.
Although these prior-driven approaches can achieve clear results to some extent, they usually involve solving optimization problems that suffer from non-convexity issues and are time-consuming.

\begin{figure*}[!t]
\vspace{-5mm}
\centering
\includegraphics[width=1\linewidth]{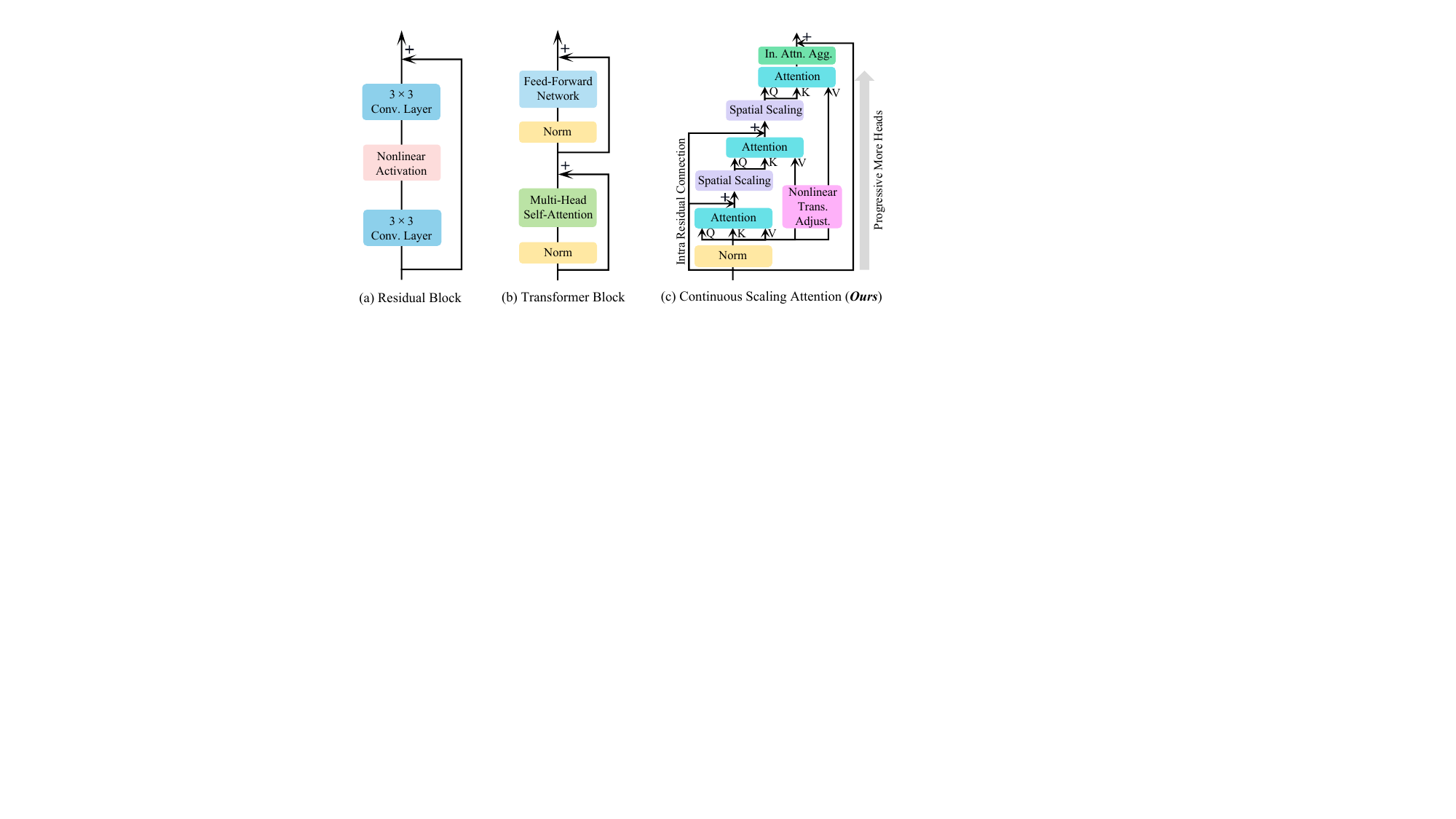}
 \vspace{-7mm}
\caption{\textbf{Comparisons on Different Modern Blocks.} \textbf{(a)} Residual block or its varieties are used to build the CNN-based image restoration methods~\cite{zhang2020rdn,SRResNet}.
\textbf{(b)} Transformer block that usually contains multi-head self-attention and a feed-forward network provide the basic for Transformer-based image restoration approaches~\cite{liang2021swinir,Zamir2021Restormer}.
\textbf{(c)} The proposed continuous scaling attention block. 
Different from CNN-based and Transformer-based blocks, we propose a Continuous Scaling Attention (\textbf{CSAttn}) block to explore how powerful potential of attention is \textbf{\textit{without using feed-forward network}}, which is essential in existing Transformers~\cite{Zamir2021Restormer,wang2021uformer,Tsai2022Stripformer,Kong_2023_CVPR_fftformer}.
We demonstrate that by inserting several key designs into continuous attention, the attention is able to achieve competitive performance compared with CNN-based and Transformer-based approaches. 
}
\label{fig: block comparisons}
\vspace{-5mm}
\end{figure*}

With the great success of convolutional neural networks (CNNs)~\cite{AlexNet,vgg19,he2016deep}, CNNs have been applied to many vision tasks and achieved remarkable performance.
CNNs have also been explored in image restoration~\cite{dehazing_AOD,ren2019progressive,liu2018desnownet,wang_aaai22,dcsfn-wang-mm20,mm20_wang_jdnet}, and outperformed previous prior-based methods. 
These CNN-based approaches are usually built on the residual block~\cite{he2016deep} or its varieties~\cite{zhang2020rdn,hu2019squeeze}, as shown in Fig.~\ref{fig: block comparisons}(a).
Due to convolution being computed in local windows and exhibiting translation equivariance, the CNN-based approaches with limited receptive field prevent it from modeling long-range pixel dependencies.

Transformers~\cite{vision_transformer,khan2021transformers} calculate the response at a given pixel by a weighted sum of all other positions instead of using local receptive windows. They have been successfully applied to image restoration tasks and achieved impressive performance~\cite{wang2021uformer,guo2022dehamer,chen2022msp,wang2023selfpromer}.
In Fig.~\ref{fig: block comparisons}(b), existing Transformer frameworks usually involve two key designs: multi-head self-attention and feed-forward network (FFN).
The former is used to perceive the long-range pixel dependencies while the latter is used for nonlinear transformation and enhancement of the features from attention.
Moreover, some studies suggest that FFNs are essential and serve as the key-value memories~\cite{geva2020transformer}.
% , which is subsequently refined throughout the model's layers via residual connections to produce the final output distribution.
%
While all the Transformers use the FFN, we would like to raise a challenge: is it flexible to use only the attention in Transformers without using FFN?
\textit{\textbf{If so, how do we design the attention to ensure that it has powerful abilities} compared with original Transformer blocks?}

In this paper, we conduct an empirical study to explore the potential of attention on image restoration.
Different from existing Transformer blocks that contain FFN after the attention computation, we propose Continuous Scaling Attention (CSAttn) which contains three attention \textit{\textbf{without using FFN}} by introducing several key designs.
\textbf{First}, instead of simply stacking 3 attention blocks, we propose Continuous Attention Learning to scale up model performance with a series of effective designs, which will be detailedly introduced below.
% Our which contains 3 continuous scaling attention to make attention empower stronger representative learning ability.
%
\textbf{Second}, different from existing attention~\cite{liu2021swin,Zamir2021Restormer} which usually does not contain the Nonlinear Activation function in attention modeling, we insert the Nonlinear Activation function into attention to activate more useful features.
\textbf{Third}, we also suggest a \textit{Value} Nonlinear Transformation Adjustment to adaptively adjust the \textit{Value} features to produce more representative information to participate in the second and third attention computation. 
\textbf{Forth}, we introduce the Intra Attention Aggregation to fuse attention features to learn better attention representation from different levels.
\textbf{Fifth}, we introduce Intra Progressive More Heads into these attentions to implicitly improve the attention representations.
\textbf{Sixth}, we suggest the Intra Residual Connections to provide more useful features into the next attention computation so that better improves restoration quality.
\textbf{At last}, we propose Spatial Scaling Learning to save the training budget while keeping the superior performance compared with the model without using scaling.
By integrating these key designs into one entity, CSAttn is able to achieve superior performance compared with CNN-based and Transformer-based image restoration approaches.
Fig.~\ref{fig: block comparisons}(c) shows the sketch of our CSAttn.

The main contributions are summarized below:
% \begin{itemize}
\begin{itemize}
     \item We propose a Continuous Scaling Attention (\textbf{CSAttn}) block to explore the potential of attention without using FFN to empower attention with strong learning ability by designing a series of key components.
    \item We analyze and show how each part affects the final restoration performance to deliver deeper insights into our proposed CSAttn.

    \item We conduct extensive experiments to demonstrate that our CSAttn favors against state-of-the-art CNN-based and Transformer-based approaches on image restoration tasks including image deraining, image desnowing, low-light image enhancement, and real image dehazing.
\end{itemize}

\vspace{-3mm}
\section{Related Work}
\vspace{-1mm}
As discussed above, current image restoration techniques can be driven by convolutional neural networks and Transformers.
\vspace{-3mm}
\subsection{CNN-based Image Restoration}
\vspace{-1mm}
With the great success of deep learning on various vision tasks, CNNs also achieve remarkable performance on image restoration~\cite{Zamir_2021_CVPR_mprnet,dudhane2021burst,zamir2020mirnet,wang_tcsvt22,icme2020-zhu} due to powerful ability of implicitly learning the priors from large-scale data.
Zhang~\etal.~\cite{DnCNN} propose a CNN-based framework to solve the image denoising problem.
Instead of directly learning the latent clear noise-free images, they devise a residual learning method to learn the noise, which achieves impressive performance compared to all previous optimization-based denoising approaches.
Pan~\etal.~\cite{dualcnn_cvpr18} suggest a dual convolutional network for low-level vision, where they formulate the convolutional networks within the task-specific formulation.
Ledig~\etal.~\cite{SRGAN} explore the generative generative adversarial networks~\cite{gan} into image super-resolution for realistic quality.
Cho~\etal.~\cite{cho2021rethinking_mimo} rethink the image deblurring task from the coarse-to-fine mechanism, where they restore the latent clear images from coarser scales to finer scales.
Zamir~\etal.~\cite{Zamir_2021_CVPR_mprnet} present a multi-stage progressive image restoration network, where they gradually recover the latent clear images from the perspective of multi-patches to entire images.
In \cite{cui2023selective},  Cui~\etal. propose a CNN-based selective frequency approach for image restoration and achieve impressive performance.
We refer the readers to recent excellent literature reviews on image restoration~\cite{anwar2019deep,li2019single}, which summarise the main designs in deep image restoration models. 
\vspace{-3mm}
\subsection{Transformer-based Image Restoration}
\vspace{-1mm}
With the attention design of Transformers~\cite{vision_transformer,khan2021transformers} which calculates responses at a given pixel by a weighted sum of all other positions instead of local receptive windows, Transformers-based approaches have been successfully applied to image restoration tasks and achieved impressive performance~\cite{wang2021uformer,guo2022dehamer,chen2022msp}.
Liang~\etal.~\cite{liang2021swinir} develop an Swin Transformer~\cite{liu2021swin} framework for image restoration.
They advance previous CNN-based image restoration approaches on various applications.
Guo~\etal.~\cite{guo2022dehamer} suggest a Transformer with transmission-aware 3D position embedding for image dehazing, where transmission implies the haze density that can provide the relative position information and haze density information to improve dehazing performance.
Zamir~\etal.~\cite{Zamir2021Restormer} introduce a Restormer to solving high-resolution image restoration problems, where they compute cross-covariance across feature channels to avoid quadratic computational complexity.
Kong~\etal.~\cite{Kong_2023_CVPR_fftformer} propose a frequency domain-based Transformer for image deblurring, where they integrate the frequency knowledge by Fast Fourier Transformation into Transformer architecture, effectively improving the deblurring results.

Different from existing Transformer architectures which usually use FFN after the attention, we propose continuous scaling attention without using FFN and achieve competitive performance on image restoration tasks by empirically designing a series of simple and effective techniques.
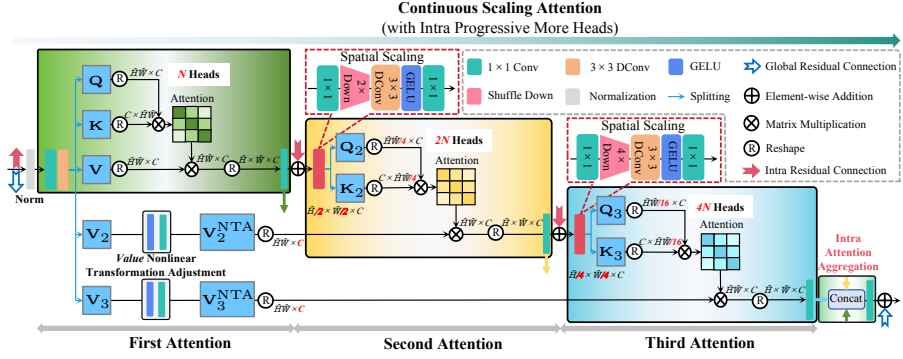
\begin{figure*}[!t]
\vspace{-3mm}
\centering
\includegraphics[width=1\linewidth]{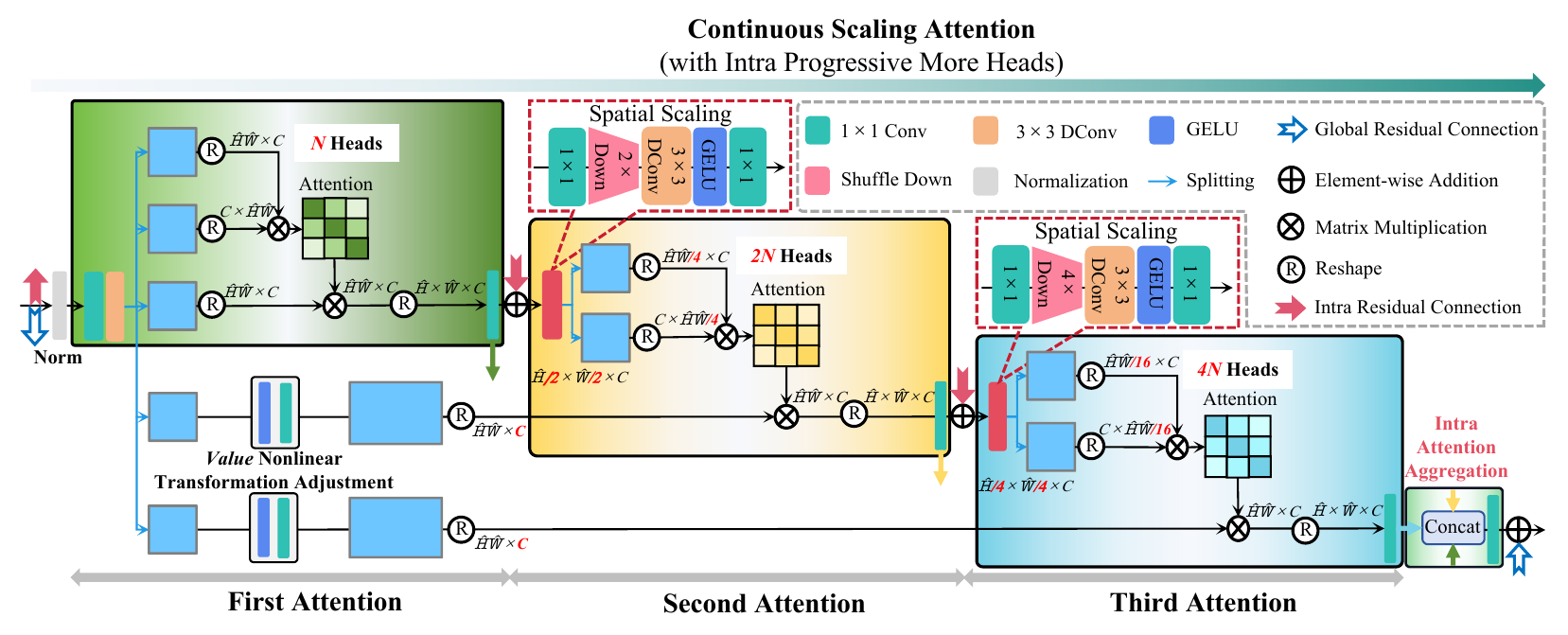}
\put(-312,105){\tiny{$\textbf{Q}$}}
\put(-312,88){\tiny{$\textbf{K}$}}
\put(-312,71){\tiny{$\textbf{V}$}}
\put(-313,47){\tiny{$\textbf{V}_{2}$}}
\put(-313,22){\tiny{$\textbf{V}_{3}$}}
\put(-269,47.5){\tiny{$\textbf{V}^{\text{NTA}}_{2}$}}
\put(-269,22){\tiny{$\textbf{V}^{\text{NTA}}_{3}$}}
\put(-217,80){\tiny{$\textbf{Q}_{2}$}}
\put(-217,64){\tiny{$\textbf{K}_{2}$}}
\put(-119,56){\tiny{$\textbf{Q}_{3}$}}
\put(-119,40){\tiny{$\textbf{K}_{3}$}}
\vspace{-3mm}
\caption{\textbf{Detailed structure of our Continuous Scaling Attention (CSAttn).
}
Our CSAttn contains three continuous attention \textit{\textbf{without using FFN}} which is essential in modern Transformer architectures~\cite{Zamir2021Restormer,wang2021uformer,vision_transformer}.
By integrating a series of simple and effective designs into one entity, CSAttn is able to achieve competitive performance compared with CNN-based and Transformer-based image restoration approaches.
}
\label{fig: Continuous Scaling Attention}
\vspace{-3mm}
\end{figure*}

\vspace{-3mm}
\section{Proposed Approach}
\vspace{-1mm}
Our goal aims to explore how powerful potential of attention on image restoration and enhancement without using a feed-forward network (FFN) which is essential in existing Transformers~\cite{Zamir2021Restormer,wang2021uformer,Tsai2022Stripformer}.
To that end, we develop the new Continuous Scaling Attention (\textbf{CSAttn}), as shown in Fig.~\ref{fig: Continuous Scaling Attention}.
Our CSAttn contains continuous three attentions without FFN.
To scale up the attention capacity, we introduce a series of key designs, including \textbf{(a)} Continuous Attention Learning; \textbf{(b)} Nonlinear Activation function; \textbf{(c)} \textit{Value} Nonlinear Transformation Adjustment; \textbf{(d)} Intra Attention Aggregation; \textbf{(e)} Intra Progressive More Heads; \textbf{(f)} Intra Residual Connections; \textbf{(g)} Spatial Scaling Learning.
Next, we will formulate our proposed CSAttn and detailedly introduce the effect of each component.
\vspace{-3mm}
\subsection{Overall pipeline}\label{sec:Overall Pipeline}
\vspace{-1mm}
Our overall network is the same as the network structure of SFNet~\cite{cui2023selective}, which is a three-level encoder and decoder framework with multiple inputs and multiple outputs.
We replace the residual blocks in SFNet~\cite{cui2023selective} with our proposed CSAttn blocks and adjust the number of blocks.
We first input 1-scale, 1/2-scale, and 1/4 scale input images to 1-3 level encoder and then reconstruct the 1-scale, 1/2-scale, and 1/4 scale output results by decoder.
We use the $L_{1}$ loss and frequency loss~\cite{cho2021rethinking_mimo} to constrain these three outputs with default parameters in \cite{cui2023selective}.

\vspace{-3mm}
\subsection{Continuous Scaling Attention}\label{sec: Continuous Scaling Attention}
\vspace{-1mm}
Our proposed Continuous Scaling Attention (CSAttn) is illustrated in Fig.~\ref{fig: Continuous Scaling Attention}, which contains three continuous attention computations with spatial scaling.
At the first attention, our CSAttn is the same as the attention in Restormer~\cite{Zamir2021Restormer}.
In the second and third attention computations, instead of stacking the same attention with the first one which may be less effective, we propose a series of simple and effective techniques to scale up the attention capacity for better model performance.

Specifically, our first attention in CSAttn can be formulated as follows:
\begin{equation}\label{eq: first}
\begin{split}
&\textbf{Q}, \textbf{K}, \textbf{V}, \textbf{V}_{2}, \textbf{V}_{3} = \mathtt{Split}\Big(W_{d}W_{p}(\textrm{$\hat{\textbf{X}}$})\Big),
\\
&\textrm{$\textbf{X}_{\textrm{first}}$} =  \mathtt{Attention}\bigg
(\textbf{Q}, \textbf{K}, \textbf{V}
\bigg),
% & \textbf{V}_{2}^{} = \mathtt{Scaling}\left(\textbf{V}_{2}\right)
\end{split}
\end{equation}
where $\mathtt{Attention}\left(\textbf{Q}, \textbf{K}, \textbf{V}\right) = \textbf{V}\cdot \textrm{Softmax$\left( \textbf{K} \cdot \textbf{Q}/\alpha \right)$}$; 
Here, $\alpha$ is a learnable scaling parameter to control the magnitude of the dot product of $\mathbf{K}$ and $\mathbf{Q}$;
$\mathtt{Split}(\cdot)$ denotes the split operation;
$W_{p}(\cdot)$ refers to the $1$$\times$$1$ point-wise convolution, while $W_{d}(\cdot)$ means the $3$$\times$$3$ depth-wise convolution.
We use $N$ heads in the first attention;
$ \textbf{V}_{2}$ and $\textbf{V}_{3}$ denote the \textit{Values} which will participate in the computation of the second and third attention;
$\textrm{$\hat{\textbf{X}}$}$ denotes the input features which are first normalized by layer normalization~\cite{ba2016layer}.

After conducting the first attention, we then compute the second and third attention.
These two attentions have similar processes with several key techniques but are different from the first attention.
The second attention, which contains three key techniques, including \textit{Value} Nonlinear Transformation Adjustment, Spatial Scaling, and Intra Residual Connection: 
\begin{equation}\label{eq: second}
\begin{split}
&\textbf{V}_{2}^{\text{NTA}} = \mathtt{\textbf{NTA}}\left(\textbf{V}_{2}\right),
\\
&\textbf{Q}_{2}, \textbf{K}_{2} = \mathtt{Split}\Big(\mathtt{\textbf{Scaling}}\left(\textrm{$\textbf{X}_{\textrm{first}}$} \oplus \textrm{$\hat{\textbf{X}}$}\right)\Big),
\\
&\textrm{$\textbf{X}_{\textrm{second}}$} =  \mathtt{Attention}\bigg
(\textbf{Q}_{2}, \textbf{K}_{2}, \textbf{V}_{2}^{\text{NTA}}
\bigg),
\end{split}
\end{equation}
where $\mathtt{\textbf{NTA}}$ denotes the \textit{Value} Nonlinear Transformation Adjustment operation, which contains a simple $1 \times 1$ convolution followed with a nonlinear activation function;
$\mathtt{\textbf{Scaling}}$ means the spatial scaling operation, which contains $1 \times 1$ point-wise convolution, suffule down operation, $3 \times 3$ depth-wise convolution, nonlinear activation function, and $1 \times 1$ point-wise convolution, where we use 2$\times$ shuffle down in the second attention;
$\oplus$ refers to the intra residual connection by adding to the original input $\textrm{$\hat{\textbf{X}}$}$.
Note that we use 2$N$ heads in the second attention.

The third attention has similar processes to the second one:
\begin{equation}\label{eq: second}
\begin{split}
&\textbf{V}_{3}^{\text{NTA}} = \mathtt{\textbf{NTA}}\left(\textbf{V}_{3}\right),
\\
&\textbf{Q}_{3}, \textbf{K}_{3} = \mathtt{Split}\Big(\mathtt{\textbf{Scaling}}\left(\textrm{$\textbf{X}_{\textrm{second}}$} \oplus \textrm{$\hat{\textbf{X}}$} \right)\Big),
\\
&\textrm{$\textbf{X}_{\textrm{third}}$} =  \mathtt{Attention}\bigg
(\textbf{Q}_{3}, \textbf{K}_{3}, \textbf{V}_{3}^{\text{NTA}}
\bigg),
\end{split}
\end{equation}
where we use 4$\times$ shuffle down in the $\mathtt{\textbf{Scaling}}$ operation and $2$$N$ heads in the third attention.

Last, we aggregate these three attention outputs to learn more representative features by concatenation and 1$\times$1 point-wise convolution, which is further added to original input $\textrm{$\hat{\textbf{X}}$}$:
\begin{equation}\label{eq: second}
\begin{split}
\textrm{$\textbf{Y}$} =  W_{p}\Big(\mathtt{Concat}\left[\textrm{$\textbf{X}_{\textrm{first}}$}, \textrm{$\textbf{X}_{\textrm{second}}$}, \textrm{$\textbf{X}_{\textrm{third}}$} \right]\Big) + \textrm{$\hat{\textbf{X}}$},
\end{split}
\end{equation}
where $\textrm{\textbf{Y}}$ means the output of CSAttn block;
$\mathtt{Concat}[\cdot]$ denotes the concatenation operation at channel dimension;
$+$ means the global residual connections in the block.

\vspace{-3mm}
\subsection{Discussion on Continuous Scaling Attention}
\vspace{-1mm}

Our CSAttn involves a series of key designs.
Discussing them to better explain our model is necessary.

{\flushleft \bf (a) Continuous Attention Learning.}
As one single attention without FFN does not work well~\cite{geva2020transformer}, we introduce a continuous attention mechanism to scale up attention capacity.
Instead of directly stacking three attention, we propose the Continuous Scaling Attention with a series of key designs that are essential to model performance.
We will demonstrate that these components will significantly benefit model capacity.

{\flushleft \bf (b) Spatial Scaling Learning.}
We propose spatial scaling learning to save the training budget while keeping superior performance compared with the model without using scaling.
The Spatial Scaling Learning contains $1 \times 1$ point-wise convolution, suffule down operation, $3 \times 3$ depth-wise convolution, nonlinear activation function, and $1 \times 1$ point-wise convolution, where the suffule down operation is used to reduce the spatial sizes.
In the second attention, we use 2$\times$ shuffle down to reduce the size of the original input to 1/2 scale, while we use 4$\times$ shuffle down to reduce the size of original input to 1/4 scale in the third attention to efficiently compute these attentions.

{\flushleft \bf (c) \textit{Value} Nonlinear Transformation Adjustment.}
\textit{Value} Nonlinear Transformation Adjustment is an essential operation, which involves very simple operations including 1$\times$1 convolution and nonlinear activation, which aims to adaptively adjust the \textit{Value} features to activate more representative content to effectively participate in the second and third attention computation.

{\flushleft \bf (d) Nonlinear Activation Function.}
The Nonlinear Activation Function is used in the operations of Spatial Scaling and \textit{Value} Nonlinear Transformation Adjustment, which is to activate more useful features for better performance.
Note that different activation functions may affect the model performance.

{\flushleft \bf (e) Intra Attention Aggregation.}
We aggregate these three attention outputs in the final to help the model learn attention features from different levels.

{\flushleft \bf (f) Intra Progressive More Heads.}
Instead of using the same number of heads from the first attention and the third attention, we suggest using progressive more heads to help the model learn different attention with different levels, which implicitly improves the attention representation to further enhance restoration quality.

{\flushleft \bf (g) Intra Residual Connections.}
The intra residual connection is introduced to ease the learning for the second and third attention to provide more useful features for the next attention computation.

\vspace{-3mm}
\section{Experiments}
\vspace{-1mm}
In this section, we evaluate our proposed \textbf{CSAttn} for $4$ image restoration tasks: \textbf{(a)} image deraining, \textbf{(b)} image desnowing, \textbf{(c)} low-light image enhancement, and \textbf{(d)} real image dehazing. 
We train separate models for different image restoration tasks. 
\vspace{-3mm}
\subsection{Implementation Details}
\vspace{-1mm}
We train \textbf{CSAttn} using the AdamW optimizer~\cite{adamw} with the initial learning rate $5e^{-4}$ that is gradually reduced to $1e^{-7}$ with the cosine annealing~\cite{loshchilov2016sgdr}.
The training patch size is set as $256$$\times$$256$ pixels.
We replace residual blocks in \cite{cui2023selective} with our proposed CSAttn blocks, which are set as [6, 6, 12] blocks in 1-3 levels if further mentioned.
We use the same loss function in~\cite{cui2023selective} with default parameters to constrain the restored results of \textbf{CSAttn} with Ground-Truth.
\begin{table*}[!t]\scriptsize
% \vspace{-3mm}
\begin{center}
\caption{\underline{\textbf{Image deraining}} results. Our \textbf{CSAttn} advances the recent 13 state-of-the-arts on average by at least $0.41$~dB PSNR.}
\label{table:deraining}
\vspace{-3.5mm}
\setlength{\tabcolsep}{0.7pt}
\scalebox{1}{
\begin{tabular}{l| c c | c c | c c | c c | c c }
\shline
\multirow{2}{*}{\textbf{Method}} & \multicolumn{2}{c|}{\textbf{Test100}}&\multicolumn{2}{c|}{\textbf{Rain100H}}&\multicolumn{2}{c|}{\textbf{Rain100L}}&\multicolumn{2}{c|}{\textbf{Test2800}}&\multicolumn{2}{c}{\textbf{Average}}\\
& PSNR~$\textcolor{black}{\uparrow}$ & SSIM~$\textcolor{black}{\uparrow}$ & PSNR~$\textcolor{black}{\uparrow}$ & SSIM~$\textcolor{black}{\uparrow}$ & PSNR~$\textcolor{black}{\uparrow}$ & SSIM~$\textcolor{black}{\uparrow}$ & PSNR~$\textcolor{black}{\uparrow}$ & SSIM~$\textcolor{black}{\uparrow}$ & 
PSNR~$\textcolor{black}{\uparrow}$ & SSIM~$\textcolor{black}{\uparrow}$\\
\shline
DerainNet~\cite{fu2017clearing} & 22.77  & 0.810  & 14.92  & 0.592  & 27.03  & 0.884  & 24.31  & 0.861    & 22.26 & 0.787    \\
SEMI~\cite{wei2019semi} & 22.35&0.788& 16.56&0.486& 25.03&0.842& 24.43&0.782 & 22.09 & 0.725 \\
DIDMDN~\cite{zhang2018density} & 22.56&0.818& 17.35&0.524& 25.23&0.741& 28.13&0.867& 23.32 & 0.738 \\
UMRL~\cite{yasarla2019uncertainty} & 24.41&0.829& 26.01&0.832& 29.18&0.923& 29.97&0.905& 27.39 & 0.872 \\
RESCAN~\cite{li2018recurrent} & 25.00&0.835& 26.36&0.786& 29.80&0.881& 31.29&0.904& 28.11 & 0.852 \\
PreNet~\cite{ren2019progressive} & 24.81&0.851& 26.77&0.858& 32.44 & 0.950 & 31.75&0.916& 28.94  & 0.894 \\
MSPFN~\cite{mspfn2020}  & 27.50 & 0.876 & 28.66 & 0.860 & 32.40 & 0.933 & 32.82 & 0.930  & 30.35 & 0.900\\
DCSFN~\cite{dcsfn-wang-mm20}  &  27.46  & 0.887   &  28.98 & 0.887  & 34.70  &  0.961 & 30.96  & 0.903    &  30.53 &  0.910\\
MPRNet~\cite{Zamir_2021_CVPR_mprnet}  & 30.27 & 0.897 & 30.41 & 0.890 & 36.40 & 0.965 & 33.64 & 0.938  & 32.68 & 0.923 \\
SPAIR~\cite{purohit2021spatially_spair} & 30.35 & 0.909 & 30.95& 0.892 & 36.93 & 0.969 & 33.34 & 0.936  & 32.89 & 0.927\\
Uformer~\cite{wang2021uformer} &29.17 &0.880&30.06&0.884&36.34&0.966&33.36&0.935&32.23&0.916 \\
% \textbf{\xnet} & \textbf{32.02} & \qmarks & \textbf{31.48} & \qmarks & \textbf{39.04} & \qmarks & \textbf{34.21} & \qmarks & \textbf{33.22} & \qmarks & \textbf{34.00} & \qmarks \\
MAXIM-2S~\cite{maxim_Tu_2022_CVPR} & 31.17&0.922&30.81&0.903&38.06&0.977&33.80&0.943&33.46&0.936 \\
SFNet~\cite{cui2023selective} & 31.47& 0.919& 31.90 &0.908 &38.21& 0.974 &33.69 &0.937 & 33.81 &0.935\\
\shline
\textbf{CSAttn} (\textbf{\textit{Ours}})&31.57 & 0.920& 31.62 & 0.906 & 39.30 & 0.979 & 34.40 &0.947  & \textcolor{red}{\textbf{34.22}} & \textcolor{red}{\textbf{0.938}} \\
\shline
\end{tabular}}
\end{center}
\vspace{-6mm}
\end{table*}

\begin{figure*}[!t]\scriptsize
\centering
\begin{center}
\begin{tabular}{ccccccccc}
\includegraphics[width=0.1235\linewidth]{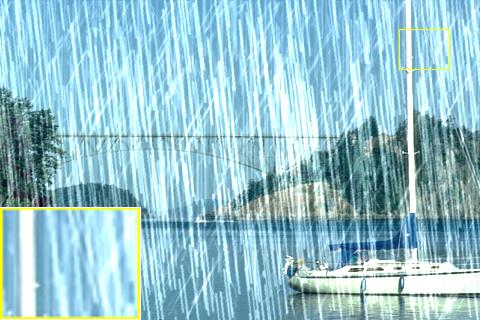} &\hspace{-1.75mm}
\includegraphics[width=0.1235\linewidth]{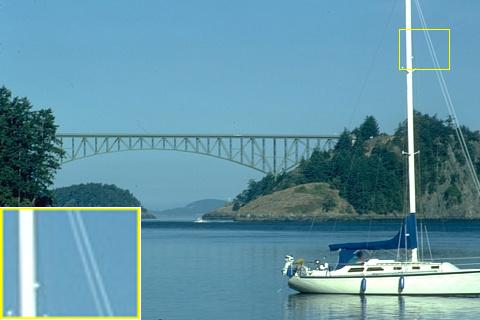} &\hspace{-1.75mm}
\includegraphics[width=0.1235\linewidth]{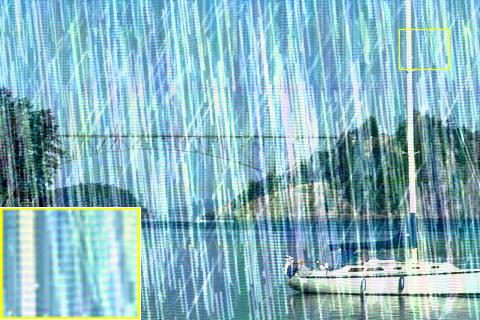}&\hspace{-1.75mm}
\includegraphics[width=0.1235\linewidth]{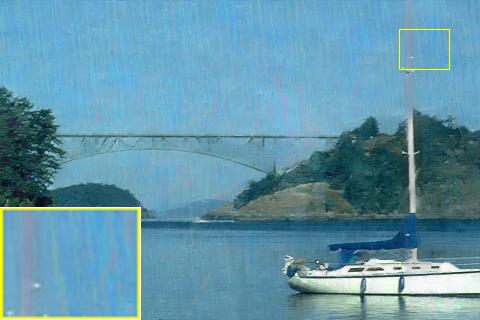} &\hspace{-1.75mm}
\includegraphics[width=0.1235\linewidth]{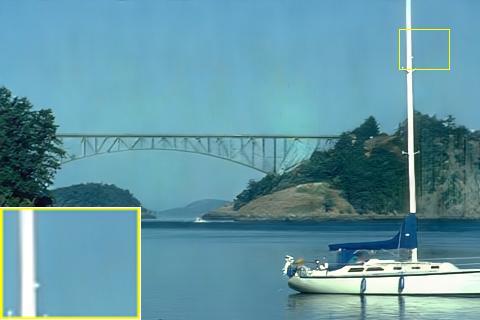} &\hspace{-1.75mm}
\includegraphics[width=0.1235\linewidth]{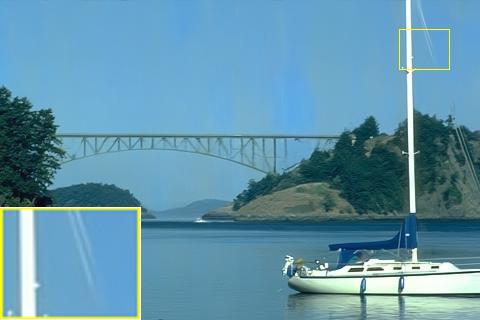} &\hspace{-1.75mm}
\includegraphics[width=0.1235\linewidth]{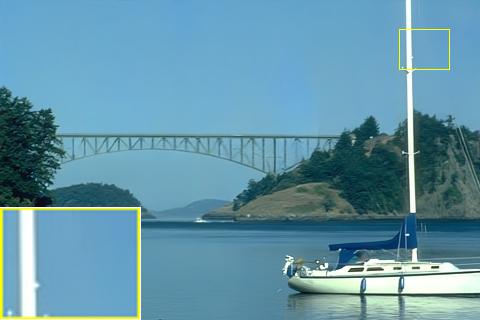} &\hspace{-1.75mm}
\includegraphics[width=0.1235\linewidth]{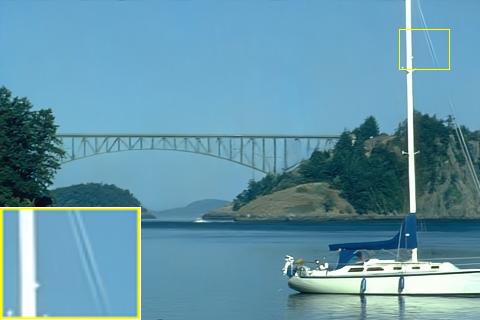}
\\
Input&\hspace{-1.75mm} GT&\hspace{-1.75mm} SEMI&\hspace{-1.75mm} RESCAN&\hspace{-1.75mm}  DCSFN&\hspace{-1.75mm} MAXIM&\hspace{-1.75mm} SFNet &\hspace{-1.75mm} \textbf{CSAttn} 
\end{tabular}
\vspace{-3mm}
\caption{\underline{\textbf{Image deraining}} on Rain100H~\cite{derain_jorder_yang}.
Our \textbf{CSAttn} is capable of restoring results with finer structures.
%
% Best viewed with zoom-in.
}
\label{fig:Image deraining example.}
\end{center}
\vspace{-8mm}
\end{figure*}

\vspace{-3mm}
\subsection{Main Results}
\vspace{-1mm}

{\flushleft \bf Results on Image Deraining.}
Following previous works~\cite{mspfn2020,Zamir_2021_CVPR_mprnet,purohit2021spatially_spair,cui2023selective}, we evaluate the image deraining task by PSNR/SSIM scores using Y channel in YCbCr color, which is trained on Rain13K dataset and then test on several individual testing datasets.
Tab.~\ref{table:deraining} summarises the quantitive results, where our CSAttn outperforms state-of-the-art approaches when averaged across these $4$ datasets, including Test100~\cite{zhang2019image}, Rain100H~\cite{yang2017deep}, Rain100L~\cite{yang2017deep}, and Test2800~\cite{fu2017clearing}. 
Compared to the best method SFNet~\cite{cui2023selective}, our CSAttn can achieve $0.41$~dB improvement on average. 
Fig.~\ref{fig:Image deraining example.} shows that our CSAttn is able to produce clearer images with finer structures, especially in the cropped region.

\begin{figure*}[!t]\scriptsize
% \vspace{-1mm}
\begin{center}
\begin{tabular}{cccccccccc}
\includegraphics[width = 0.141\linewidth]{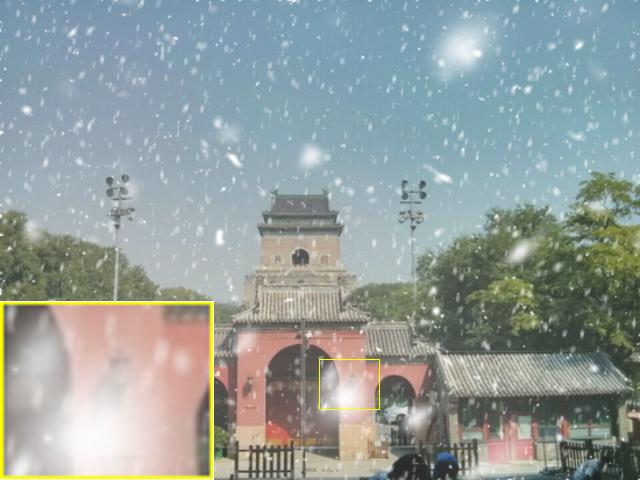}&\hspace{-1.75mm}
\includegraphics[width = 0.141\linewidth]{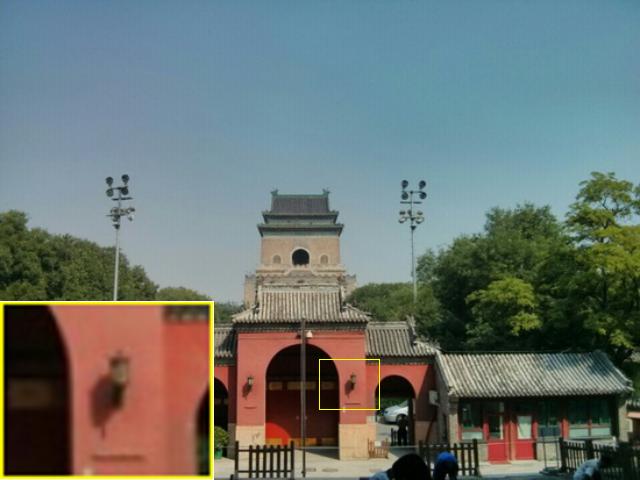}&\hspace{-1.75mm}
\includegraphics[width = 0.141\linewidth]{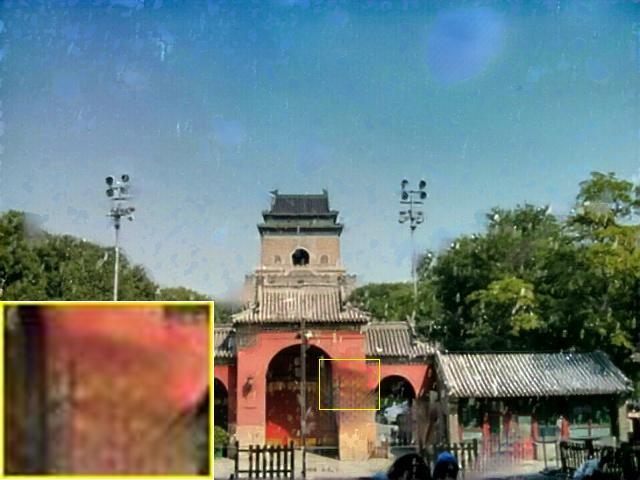}&\hspace{-1.75mm}
\includegraphics[width = 0.141\linewidth]{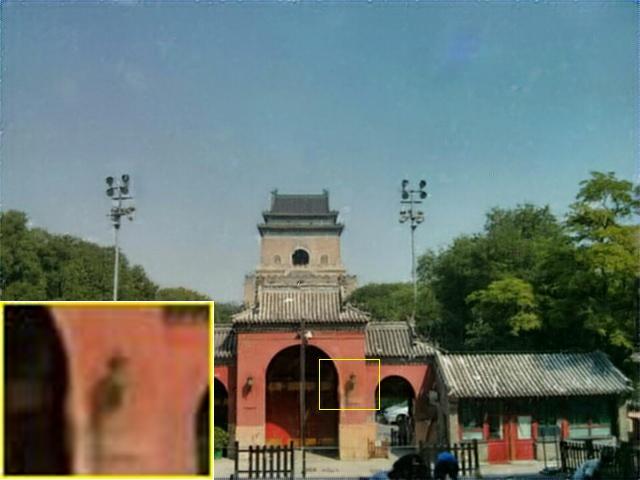}&\hspace{-1.75mm}
\includegraphics[width = 0.141\linewidth]{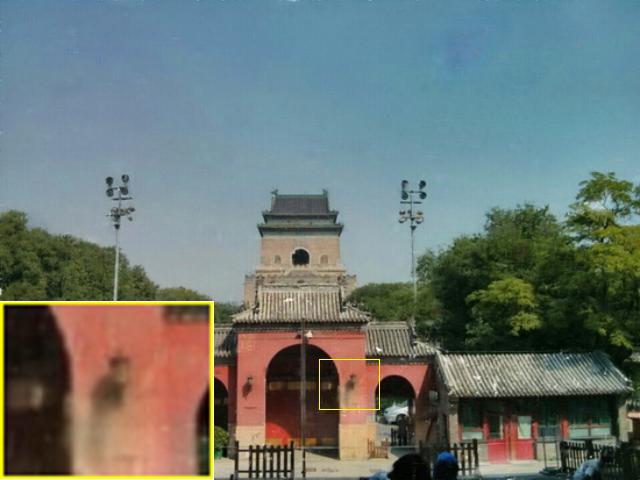}&\hspace{-1.75mm}
\includegraphics[width = 0.141\linewidth]{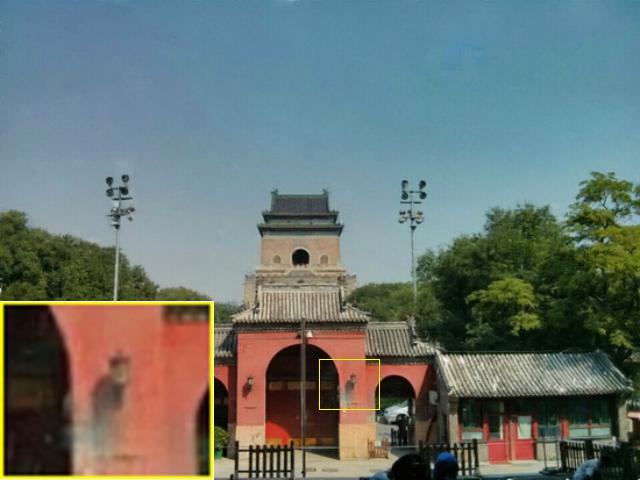}&\hspace{-1.75mm}
\includegraphics[width = 0.141\linewidth]{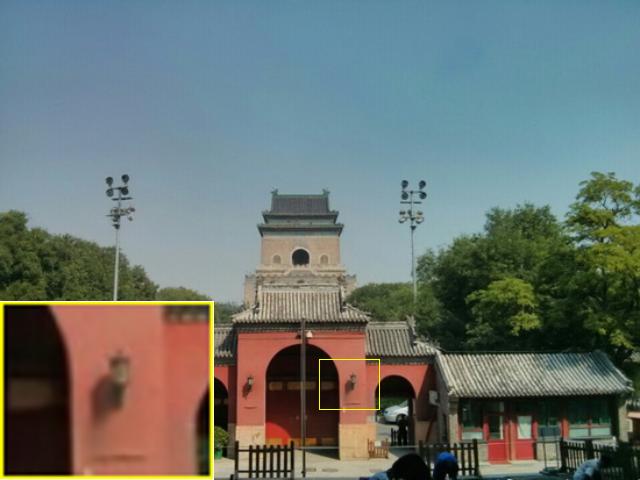}
\\
Input  &\hspace{-1.75mm}  GT   &\hspace{-1.75mm}    JSTASR  &\hspace{-1.75mm}   HDCWNet&\hspace{-1.75mm}  Uformer&\hspace{-1.75mm} Restormer  &\hspace{-1.75mm}  \textbf{CSAttn}
\\
\end{tabular}
\vspace{-3mm}
\caption{\underline{\textbf{Image desnowing}} on CSD (2000)~\cite{chen2021all}.
\textbf{CSAttn} is able to recover much cleaner results with fewer artifacts.
}
\label{fig: Visual comparisons of image desnowing on CSD (2000).}
\end{center}
\vspace{-6mm}
\end{figure*}

{\flushleft \bf Results on Image Desnowing.}
We compare our CSAttn on image desnowing task with CNN-based desnowing competitors~\cite{chen2020jstasr,chen2021all} and Transformer-based approaches~\cite{chen2022msp,valanarasu2022transweather,Zamir2021Restormer,wang2021uformer,PromptRestorer} on CSD~\cite{chen2021all} and Snow100K~\cite{liu2018desnownet} datasets.
Moreover, we also compare with Transformer-based general image restoration approaches like Restormer~\cite{Zamir2021Restormer}, Uformer~\cite{wang2021uformer} and PromptRestorer~\cite{PromptRestorer}.
The results are reported in Tab.~\ref{table:Image snowing results}, where our CSAttn outperforms state-of-the-art approach PromptRestorer~\cite{PromptRestorer} on both CSD~\cite{chen2021all} and Snow100K~\cite{liu2018desnownet} benchmarks. 
Fig.~\ref{fig: Visual comparisons of image desnowing on CSD (2000).} shows that our CSAttn removes spatially varying snows while state-of-the-art methods always hand down some artifacts.

\begin{table}[!t]\scriptsize
% \vspace{-3mm}
\begin{center}
\caption{\underline{\textbf{Image desnowing}} results on CSD (2000)~\cite{chen2021all}, and Snow100K (2000)~\cite{rim_2020_realblur}.
Our \textbf{CSAttn} achieves the best PSNR and SSIM metrics on image desnowing.
}
\label{table:Image snowing results}
\vspace{-3mm}
\setlength{\tabcolsep}{1pt}
\scalebox{0.7}{
\begin{tabular}{l | c | c  c  c cccccccccccc}
\shline
 \multirow{2}{*}{\textbf{Benchmark}} & \multirow{2}{*}{\textbf{Metrics}}&DesnowNet&JSTASR&HDCW-Net &TransWeather&MSP-Former&Uformer&Restormer~& PromptRestorer&\textbf{CSAttn}\\
  & &\cite{liu2018desnownet}&\cite{chen2020jstasr}&\cite{chen2021all} &\cite{valanarasu2022transweather}&\cite{chen2022msp}&\cite{Wang_2022_CVPR}&\cite{Zamir2021Restormer}& \cite{PromptRestorer}& \textbf{\textit{Ours}}\\
\shline
\multirow{2}{*}{\textbf{CSD (2000)}~\cite{chen2021all}}& PSNR~$\textcolor{black}{\uparrow}$  &20.13  & 27.96  &  29.06&31.76&33.75&33.80&35.43&37.48 &\textcolor{red}{\textbf{37.78}}\\
& SSIM~$\textcolor{black}{\uparrow}$     & 0.81  & 0.88  & 0.91&0.93&0.96&0.96&0.97&0.99  &\textcolor{red}{\textbf{0.99}}\\\shline
\multirow{2}{*}{\textbf{Snow100K (2000)}~\cite{liu2018desnownet}}& PSNR~$\textcolor{black}{\uparrow}$     &30.50  & 23.12&31.54& 31.82&33.43 &33.81&34.67&36.02& \textcolor{red}{\textbf{36.08}}\\
& SSIM~$\textcolor{black}{\uparrow}$     &0.94  & 0.86  &  0.95&0.95&0.96&0.94&0.95&0.97 & \textcolor{red}{\textbf{0.97}}\\
\shline
\end{tabular}}
\end{center}
\vspace{-5mm}
\end{table}
\begin{table}[!t]\scriptsize
\begin{center}
\caption{\underline{\textbf{Low-light image enhancement}} results on LOL~\cite{retinexnet_wei_bmvc18}.
Our \textbf{CSAttn} significantly advances the state-of-the-art approaches by at least $4.22$~dB PSNR.
}
\label{table:Low-light image enhancement results}
\vspace{-3mm}
\setlength{\tabcolsep}{1pt}
\scalebox{0.895}{
\begin{tabular}{l | c | c  c  c cccccccccccc}
\shline
\multirow{2}{*}{\textbf{Benchmark}} & \multirow{2}{*}{\textbf{Metrics}}&Retinex-Net&Zero-DCE&AGLLNet &Zhao~et al. & RUAS & SCI & URetinex & UHDFour & \textbf{CSAttn}
\\
& &\cite{retinexnet_wei_bmvc18}&\cite{zerodce_lowlight_guo}&\cite{AGLLNet} &\cite{zhao_lie}&\cite{RUAS_liu_cvpr21}&\cite{ma2022toward}&\cite{Wu_2022_CVPR}& \cite{Li2023ICLR_uhdfour}&\textbf{\textit{Ours}}
  \\
\shline
\multirow{2}{*}{\textbf{LOL}~\cite{chen2021all}}& PSNR~$\textcolor{black}{\uparrow}$  &16.77  & 16.79  &  17.52&21.67&16.44&14.78&19.84&23.09 &\textcolor{red}{\textbf{27.31}}\\
 & SSIM~$\textcolor{black}{\uparrow}$    & 0.54  & 0.67 & 0.77& 0.87&0.70& 0.62 &0.87&0.87  &\textcolor{red}{\textbf{0.93}}\\\shline
\end{tabular}}
\end{center}
\vspace{-5mm}
\end{table}
\begin{figure*}[!t]\scriptsize
% \vspace{-1mm}
\begin{center}
\begin{tabular}{cccccccccc}
\includegraphics[width = 0.141\linewidth]{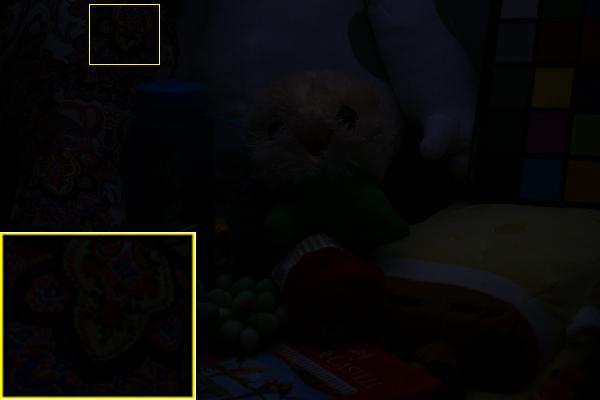}&\hspace{-1.75mm}
\includegraphics[width = 0.141\linewidth]{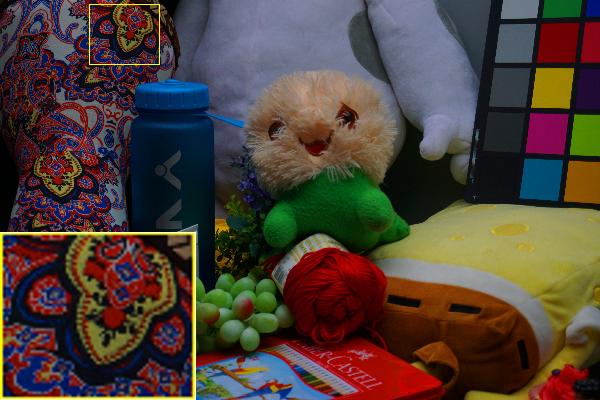}&\hspace{-1.75mm}
\includegraphics[width = 0.141\linewidth]{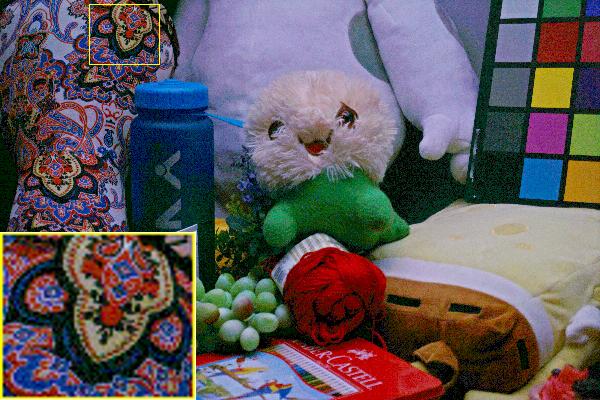}&\hspace{-1.75mm}
\includegraphics[width = 0.141\linewidth]{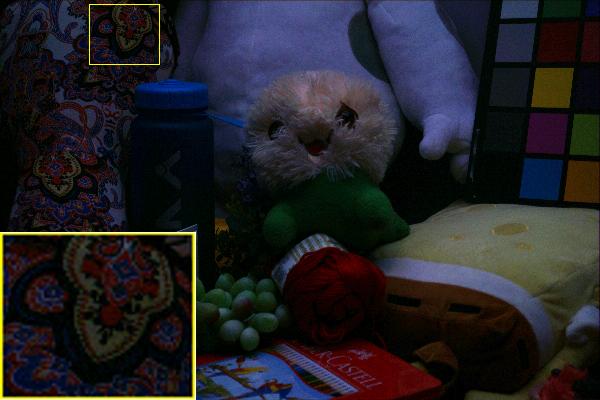}&\hspace{-1.75mm}
\includegraphics[width = 0.141\linewidth]{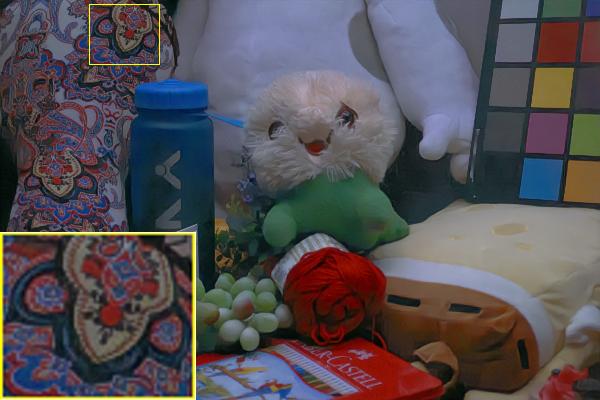}&\hspace{-1.75mm}
\includegraphics[width = 0.141\linewidth]{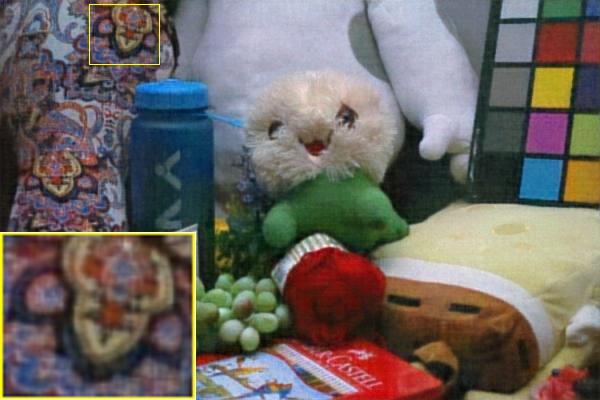}&\hspace{-1.75mm}
\includegraphics[width = 0.141\linewidth]{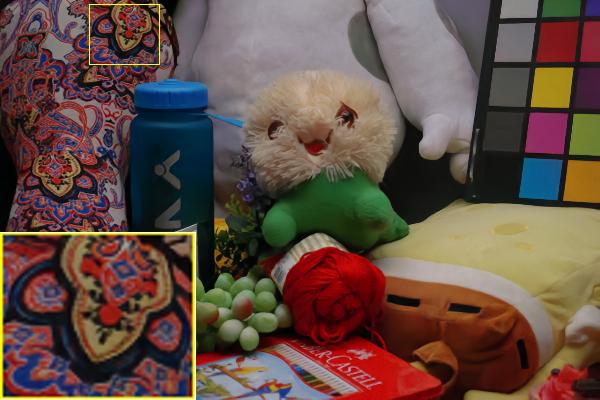}
\\
Input  &\hspace{-1.75mm} GT   &\hspace{-1.75mm}    ZeroDCE  &\hspace{-1.75mm}   SCI&\hspace{-1.75mm}    URetinex&\hspace{-1.75mm}  UHDFour  &\hspace{-1.75mm}  \textbf{CSAttn}
\\
\end{tabular}
\vspace{-3mm}
\caption{\underline{\textbf{Low-light image enhancement}} on LOL~\cite{retinexnet_wei_bmvc18}.
Our \textbf{CSAttn} is able to generate results with more natural colors.
%
% Best viewed with zoom-in.
}
\label{fig: Low-light image enhancement example on LOL.}
\end{center}
\vspace{-5mm}
\end{figure*}

{\flushleft \bf Results on Low-light Image Enhancement.}
We conduct the experiment of low-light image enhancement on widely-used LOL benchmark~\cite{retinexnet_wei_bmvc18}. 
Tab.~\ref{table:Low-light image enhancement results} reports the quantitative results.
One can obviously observe that our CSAttn achieves $4.22$~dB PSNR gains compared to recent works UHDFour~\cite{Li2023ICLR_uhdfour}. 
Fig.~\ref{fig: Low-light image enhancement example on LOL.} shows our CSAttn is able to generate more natural results with vivid colors.

\begin{figure}[!t]\scriptsize
% \vspace{-1mm}
\begin{center}
\begin{tabular}{cccccccccc}
\includegraphics[width = 0.249\linewidth]{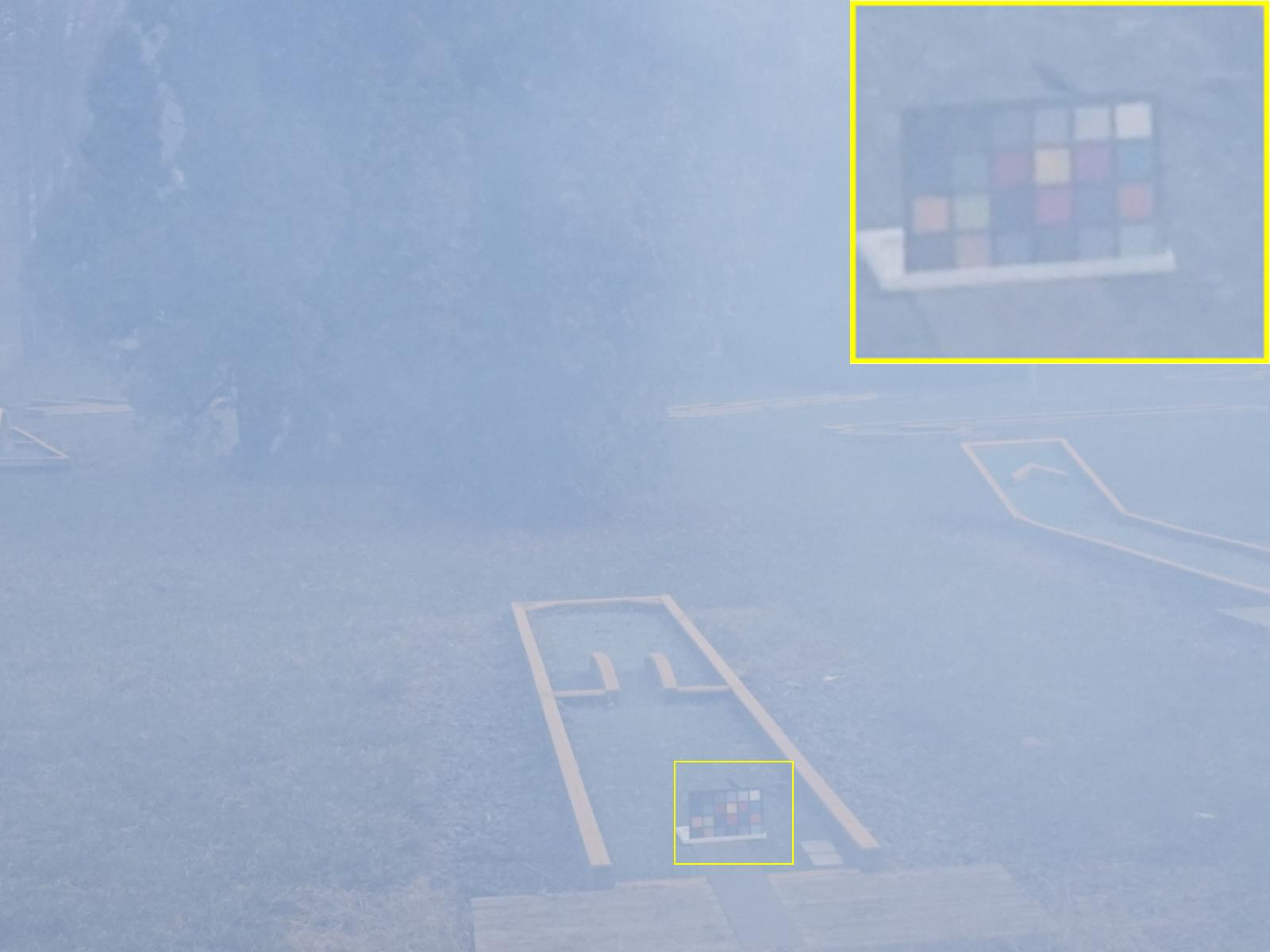}&\hspace{-1.75mm}
\includegraphics[width = 0.249\linewidth]{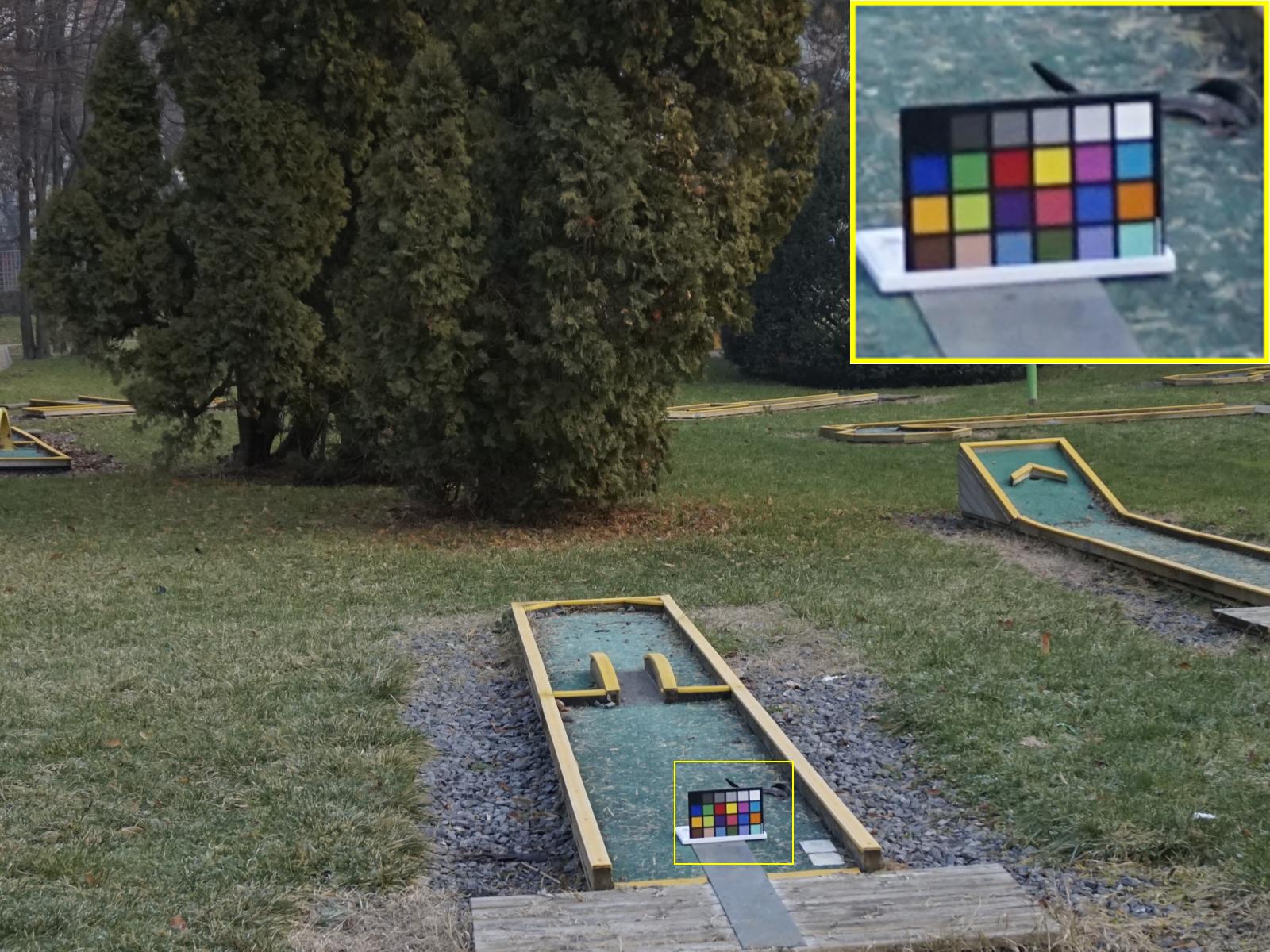}&\hspace{-1.75mm}
\includegraphics[width = 0.249\linewidth]{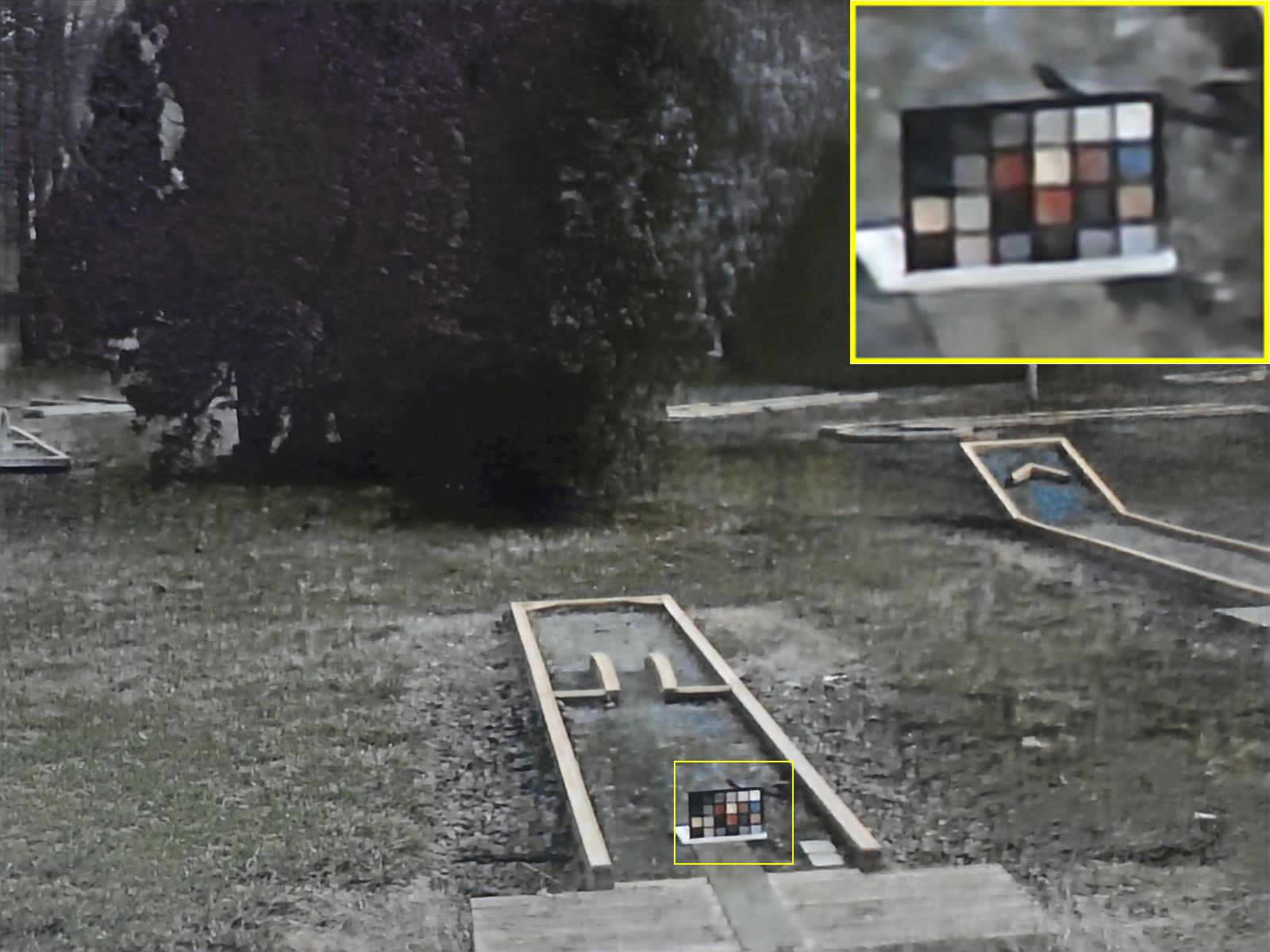}&\hspace{-1.75mm}
\includegraphics[width = 0.249\linewidth]{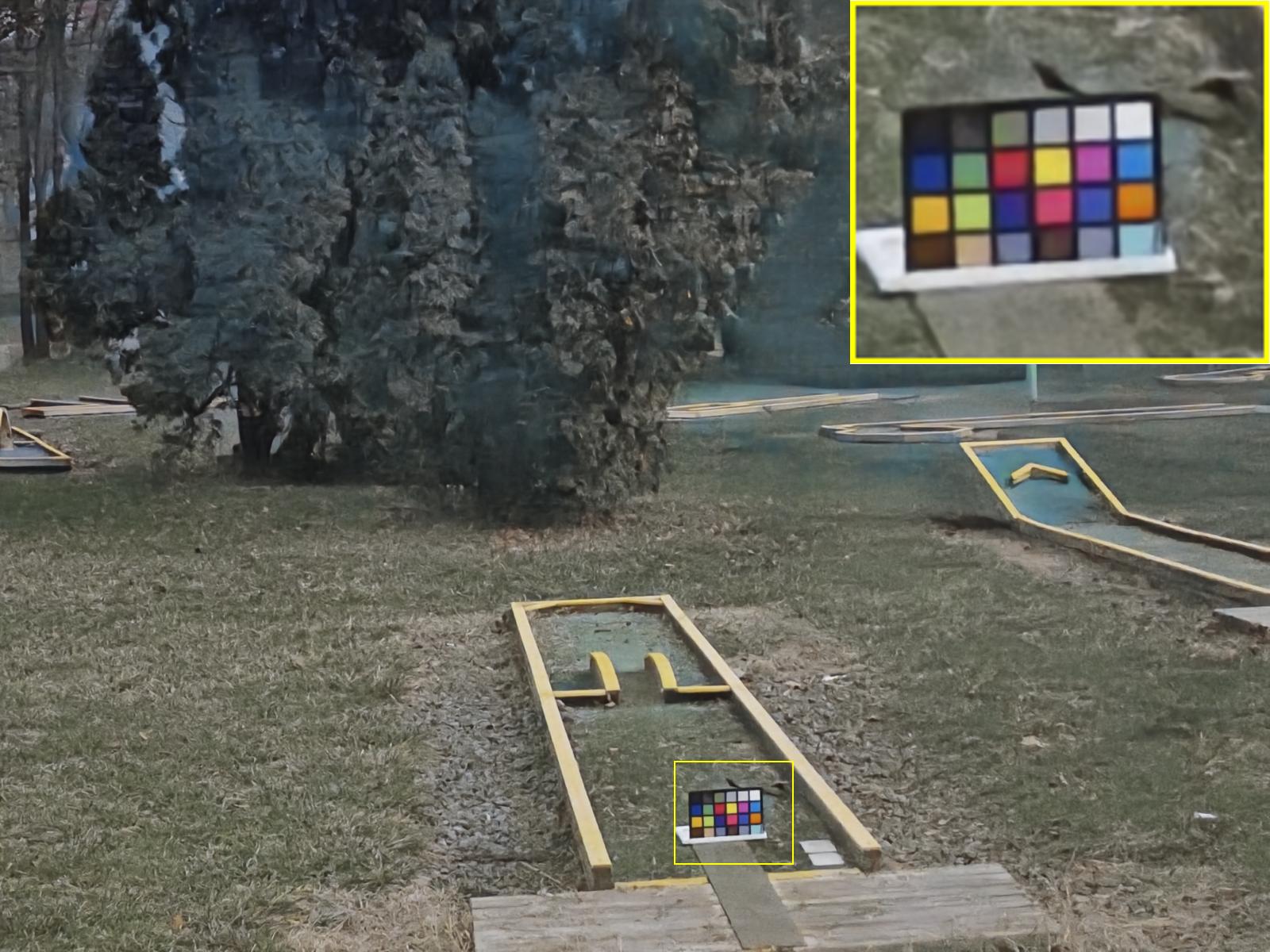}
\\
\includegraphics[width = 0.249\linewidth]{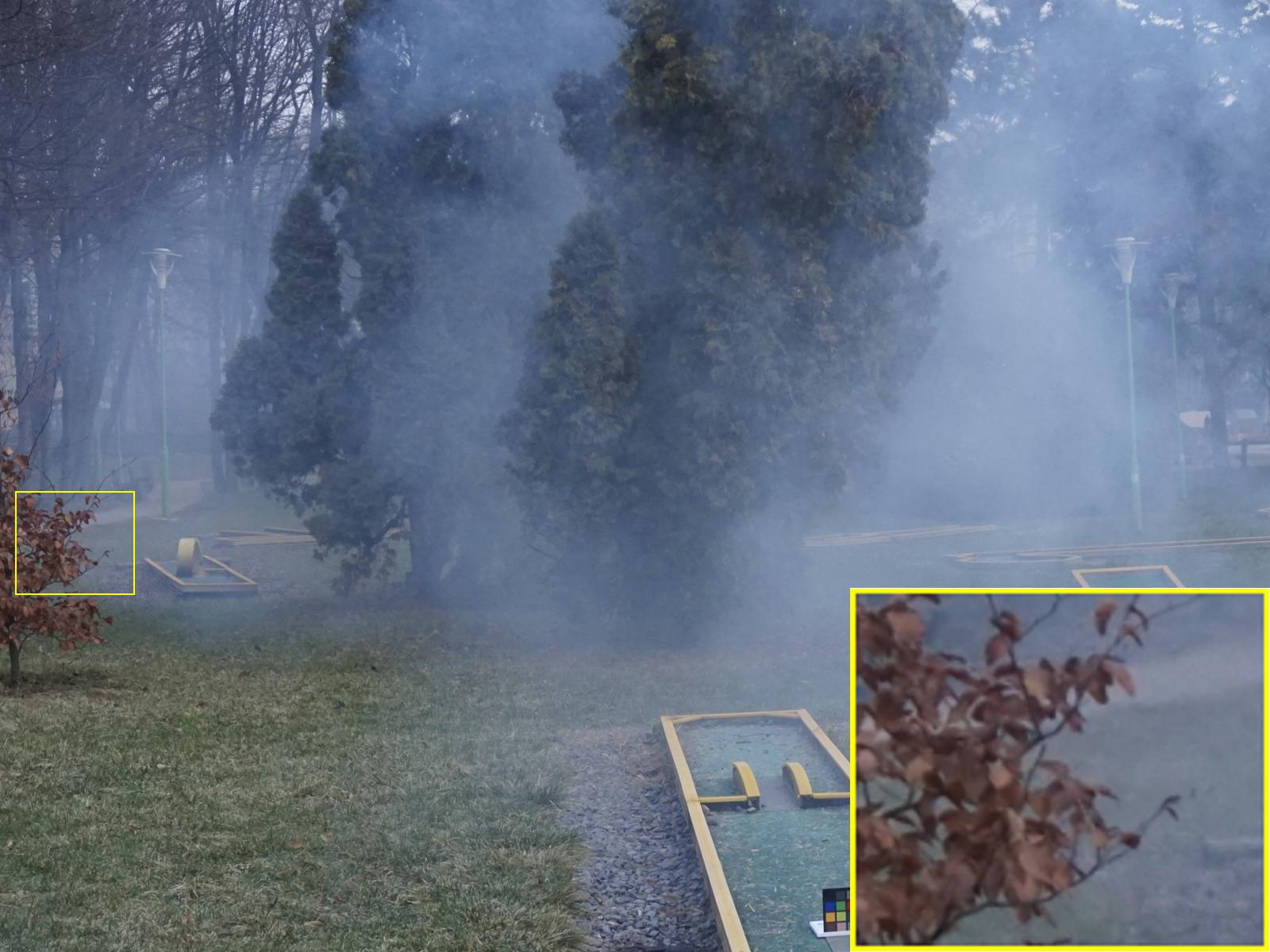}&\hspace{-1.75mm}
\includegraphics[width = 0.249\linewidth]{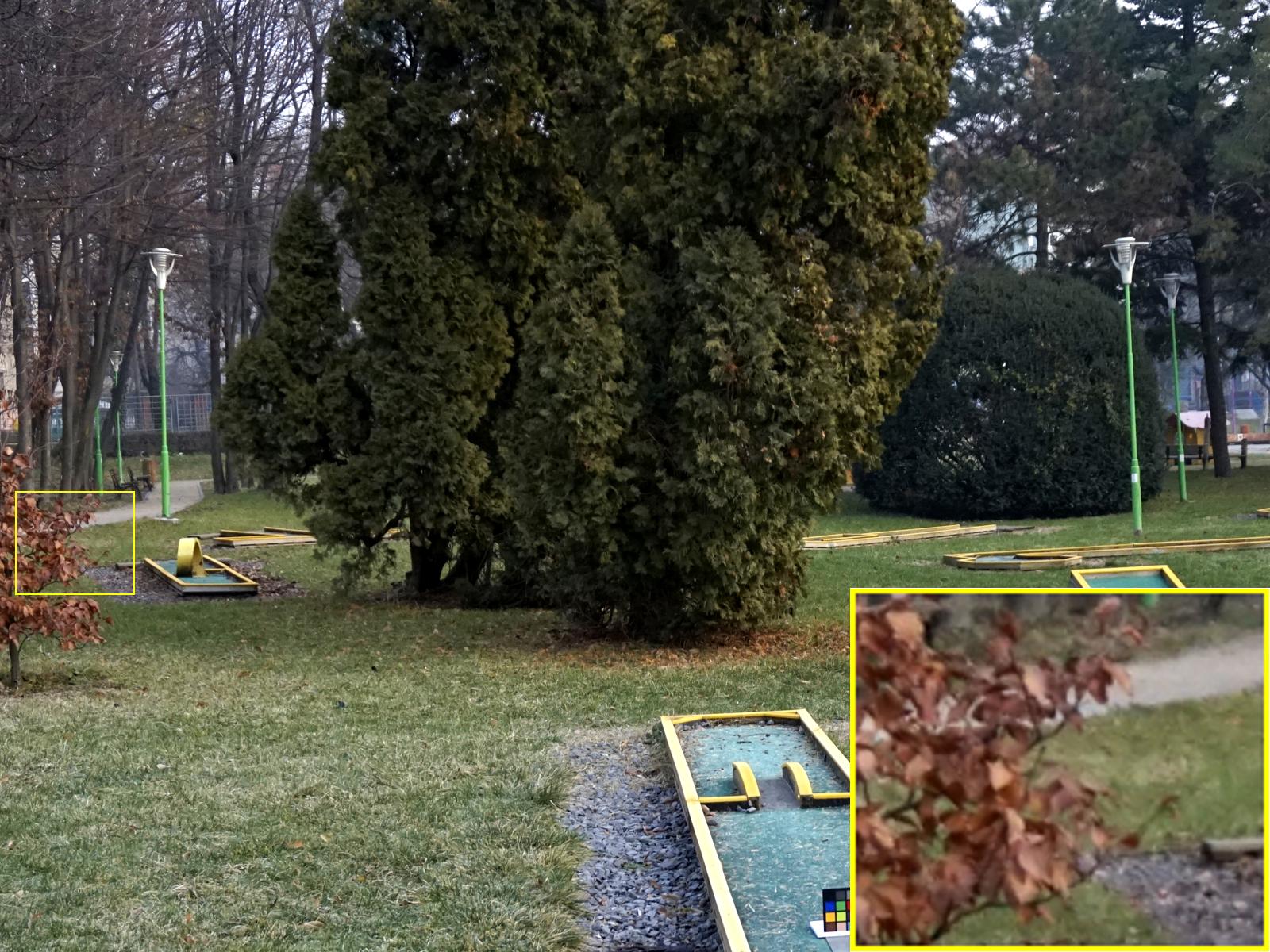}&\hspace{-1.75mm}
\includegraphics[width = 0.249\linewidth]{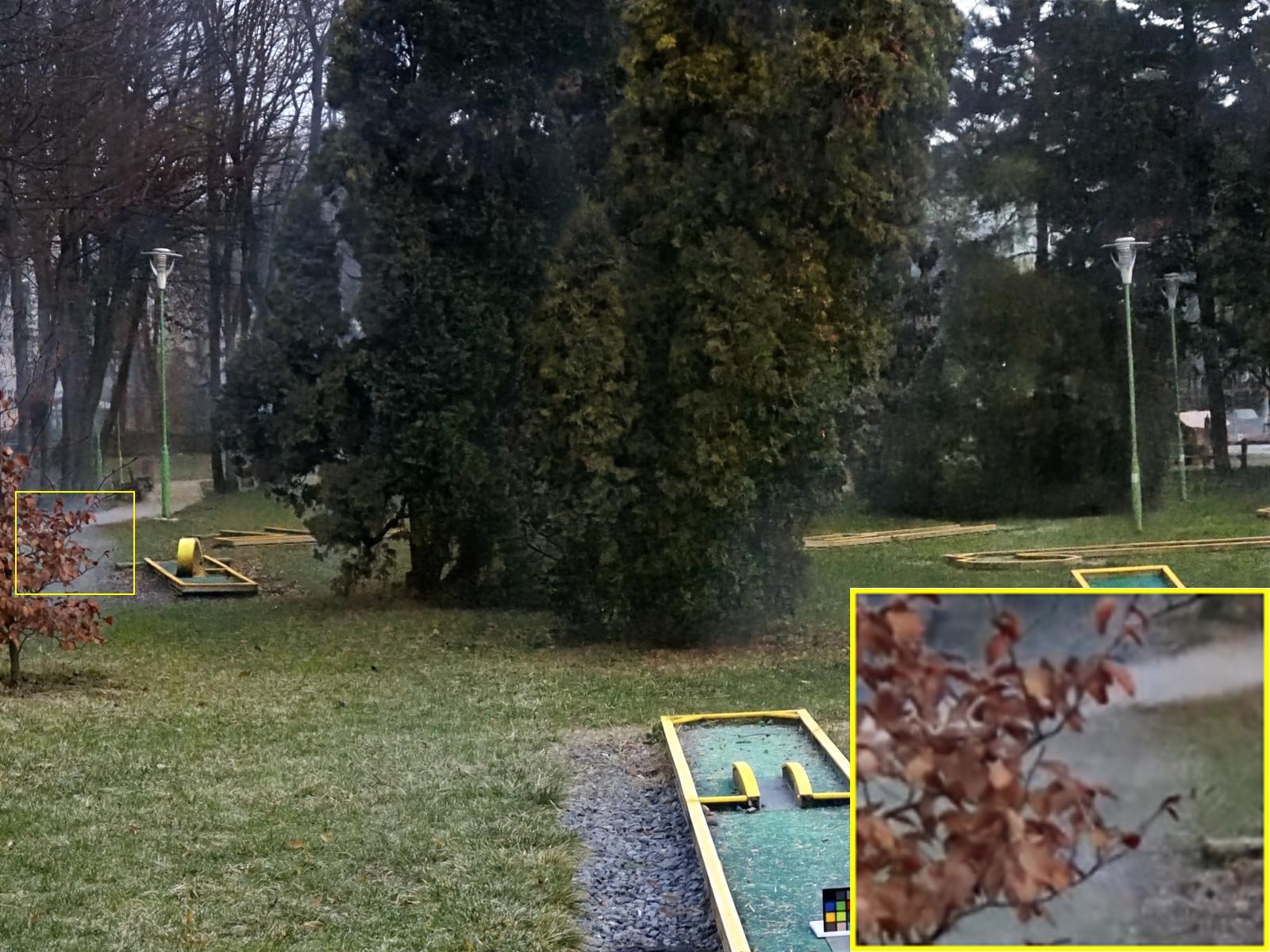}&\hspace{-1.75mm}
\includegraphics[width = 0.249\linewidth]{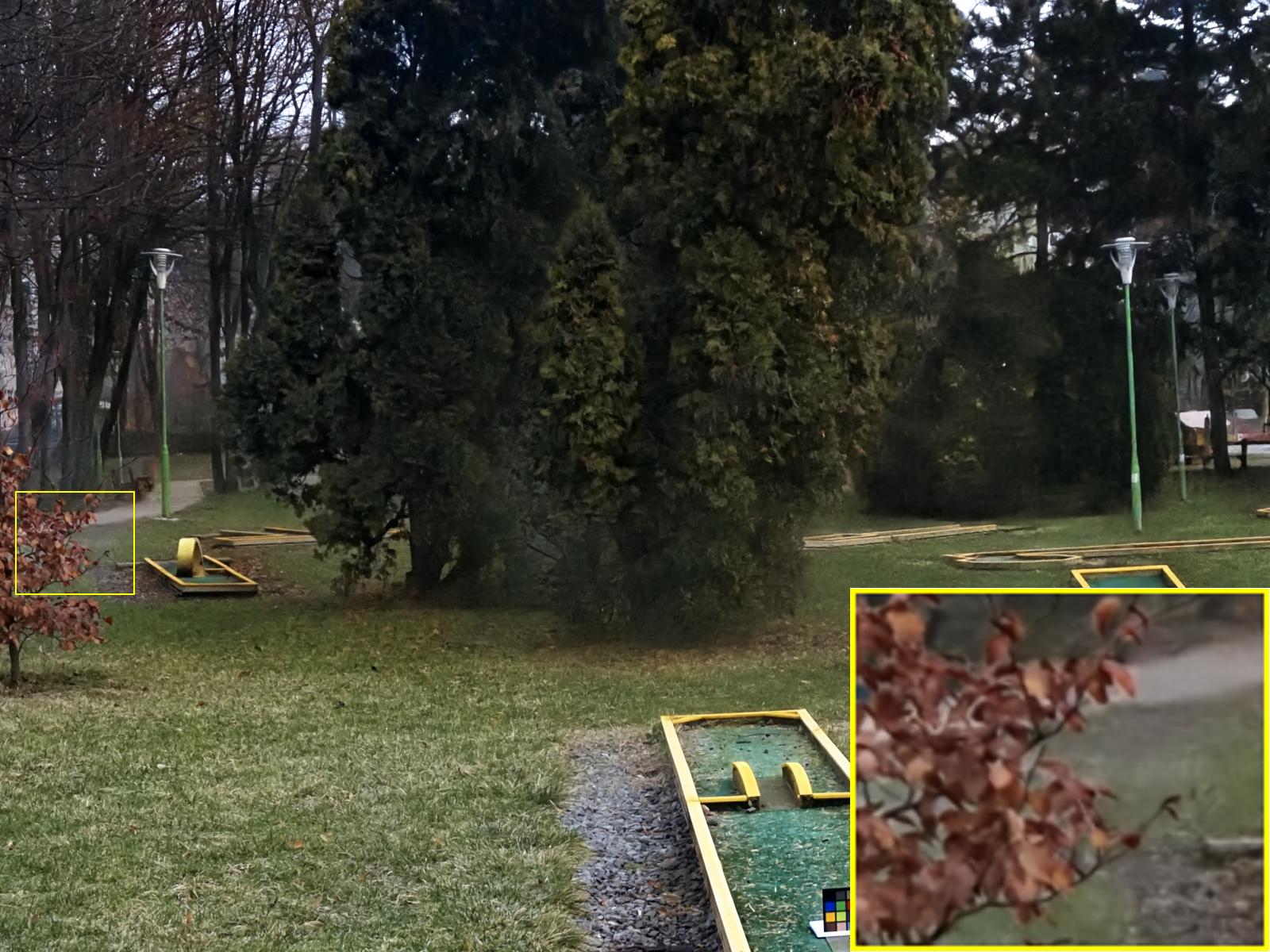}
\\
 Input  &\hspace{-1.75mm}  GT  &\hspace{-1.75mm} DeHamer  &\hspace{-1.75mm}  Ours
\end{tabular}
\vspace{-3mm}
\caption{ \textbf{Real image dehazing} on Dense-Haze~\cite{ancuti2019dense} (\textbf{Top}) and NH-Haze~\cite{ancuti2020nh} (\textbf{Bottom}).
Our \textbf{CSAttn} is capable of recovering results with less haze residuals and more natural colors.
%
% Best viewed with zoom-in.
}
\label{fig: Real image dehazing example on Dense-Haze and NH-Haze.}
\end{center}
\vspace{-8mm}
\end{figure}

{\flushleft \bf Results on Real Image Dehazing.}
We conduct real image dehazing experiments on real hazy benchmarks: Dense-Haze~\cite{ancuti2019dense} and NH-Haze~\cite{ancuti2020nh}. 
Especially, compared to recent state-of-the-art works PromptRestorer~\cite{PromptRestorer}, our CSAttn receives $6.03\%$ SSIM on Dense-Haze~\cite{ancuti2019dense}, demonstrating the effectiveness of our CSAttn by achieving significant improvement. 
Fig.~\ref{fig: Real image dehazing example on Dense-Haze and NH-Haze.} shows our CSAttn is able to generate clearer results with more vivid colors and less haze residuals.

\begin{table}[!t]\scriptsize
\vspace{-3mm}
\begin{center}
\caption{\underline{\textbf{Image dehazing}} results on real benchmarks Dense-Haze~\cite{ancuti2019dense} and NH-Haze~\cite{ancuti2020nh}.
Our \textbf{CSAttn} outperforms recent state-of-the-art method~\cite{PromptRestorer}.
}
\label{table:Image dehazing}
\vspace{-3mm}
\setlength{\tabcolsep}{1pt}
\scalebox{0.82}{
\begin{tabular}{l | c | c  c  c cccccccccccc}
\shline
 \multirow{2}{*}{\textbf{Benchmark}} & \multirow{2}{*}{\textbf{Metrics}}&DCP&DehazeNet&AOD&GridNet&FFANet &MSBDN &UHD&DeHamer & PromptRestorer& \textbf{CSAttn}\\
 & &\cite{he2010single}&\cite{dehazing_tip16_dehazenet}&\cite{dehazing_AOD} &\cite{grid_dehaze_liu}&\cite{qin2020ffa} &\cite{msbdn_cvpr20_dong} &\cite{Zheng_uhd_CVPR21}&\cite{guo2022dehamer} & \cite{PromptRestorer}& \textbf{\textit{Ours}}\\
\shline
\multirow{2}{*}{\textbf{Dense-Haze}~\cite{ancuti2019dense}}& PSNR~$\textcolor{black}{\uparrow}$  &11.01 & 9.48  &12.82&14.96&12.22&15.13&12.16 &\textcolor{red}{\textbf{16.62}}& 15.86&16.33\\
& SSIM~$\textcolor{black}{\uparrow}$     & 0.4165 &0.4383  &0.4683   &0.5326&0.4440&0.5551&0.4594&0.5602&0.5680& \textcolor{red}{\textbf{0.6283}}  \\\shline
\multirow{2}{*}{\textbf{NH-Haze}~\cite{ancuti2020nh}}& PSNR~$\textcolor{black}{\uparrow}$     &12.72  & 11.76  &  15.69&18.33&18.13&17.97&16.05&20.66& 20.36&\textcolor{red}{\textbf{20.66}}\\
& SSIM~$\textcolor{black}{\uparrow}$     & 0.4419  & 0.3988  &0.5728&0.6667&0.6473&0.6591&0.4612  &0.6844 &0.7203&\textcolor{red}{\textbf{0.7209}}  \\
\shline
\end{tabular}}
\end{center}
\vspace{-6mm}
\end{table}

\vspace{-3mm}
\subsection{Ablation Study}\label{sec:Analysis and Discussion}
\begin{figure}[!t]\scriptsize
\vspace{-3mm}
\begin{center}
\begin{tabular}{cccccccccc}
\includegraphics[width = 0.249\linewidth]{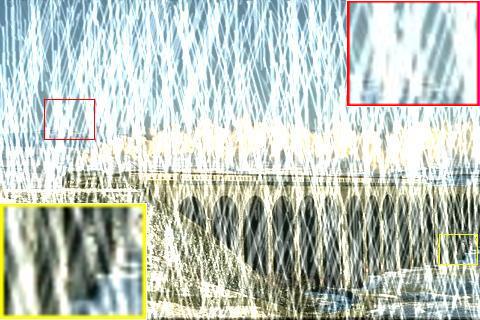}&\hspace{-1.75mm}
\includegraphics[width = 0.249\linewidth]{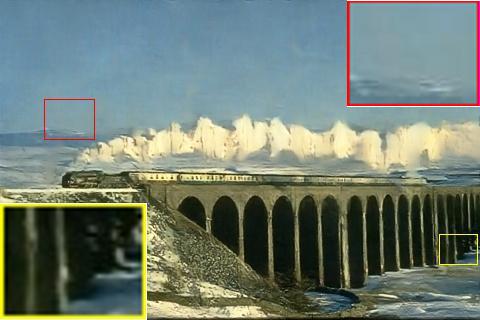}&\hspace{-1.75mm}
\includegraphics[width = 0.249\linewidth]{ablation/Rect/Rect_Rect_ablation-wovalue-3__.jpg}&\hspace{-1.75mm}
\includegraphics[width = 0.249\linewidth]{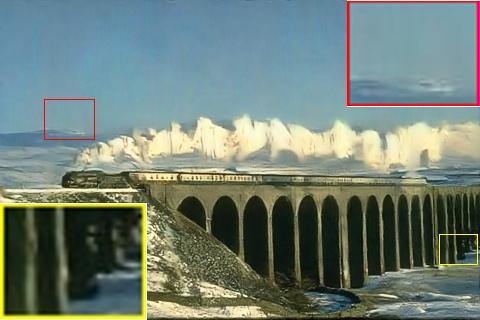}
\\
 Input  &\hspace{-1.75mm}  Tab.~\ref{table: Ablation study}(a)&\hspace{-1.75mm} Tab.~\ref{table: Ablation study}(b)&\hspace{-1.75mm}  Tab.~\ref{table: Ablation study}(c)
\\
\includegraphics[width = 0.249\linewidth]{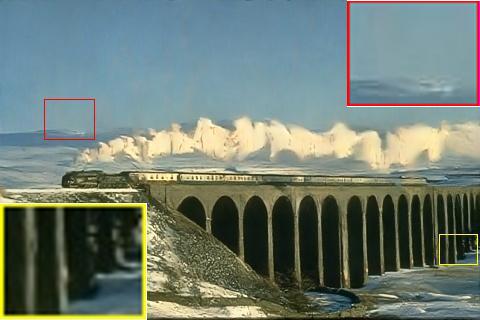}&\hspace{-1.75mm}
\includegraphics[width = 0.249\linewidth]{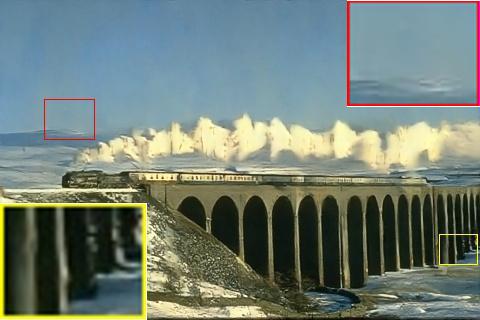}&\hspace{-1.75mm}
\includegraphics[width = 0.249\linewidth]{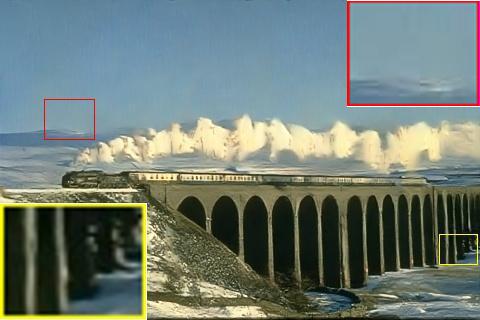}&\hspace{-1.75mm}
\includegraphics[width = 0.249\linewidth]{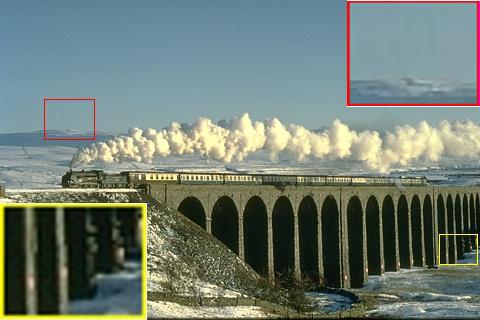}
\\
Tab.~\ref{table: Ablation study}(d)&\hspace{-1.75mm} Tab.~\ref{table: Ablation study}(e)  &\hspace{-1.75mm}  Tab.~\ref{table: Ablation study}(f) (\textbf{\textit{Ours}})  &\hspace{-1.75mm}  GT
\end{tabular}
\vspace{-3mm}
\caption{ \textbf{Visual Effect on Each Component}.
Our full model is able to produce results with better details and fewer artifacts.
%
% Best viewed with zoom-in.
}
\label{fig: ablation study.}
\end{center}
\vspace{-8mm}
\end{figure}

\vspace{-1mm}
We conduct the ablation study on the Rain100H testing dataset using the model trained on the Rain13K dataset for $100$ epochs only.
% on our tiny model version. 
%
When analyzing our model, the number of blocks is set as $[3, 6, 8]$ from the 1-3 level and the number of channels is set as 32.
Next, we describe the influence of each component individually.
Tab.~\ref{table: Ablation study} and Fig.~\ref{fig: ablation study.} respectively show the quantitative and qualitative results.

{\flushleft \bf Effect on Nonlinear Activation.}
As we use the nonlinear activation in the operations of Spatial Scaling and \textit{Value} Nonlinear Transformation Adjustment, analyzing its effectiveness is necessary.
Tab.~\ref{table: Ablation study} shows that the nonlinear activation function plays a crucial role in image restoration quality (Tab.~\ref{table: Ablation study}(a) vs. Tab.~\ref{table: Ablation study}(f)).
It can respectively improve PSNR and SSIM by $1.032$~dB and 1.58\% without consuming additional parameters and FLOPs.
Further, we note that most existing attention modules~\cite{liu2021swin,Zamir2021Restormer} do not utilize nonlinear activation.
However, our study clearly suggests that using nonlinear activation in attention is essential, which can learn more representative features for the next attention computation for better performance.

{\flushleft \bf Effect on \textit{Value} Nonlinear Transformation Adjustment.}
We use \textit{Value} Nonlinear Transformation Adjustment by simply utilizing $1 \times 1$ convolution and nonlinear activation to participate in the second and third attention computation.
One may wonder to know how it affects the restoration quality.
Tab.~\ref{table: Ablation study} shows that nonlinear transformation adjustment can improve the PSNR and SSIM by $0.883$~dB and 1.77\%, respectively (Tab.~\ref{table: Ablation study}(b) vs. Tab.~\ref{table: Ablation study}(f)).
It indicates that adaptively adjusting the \textit{Value} is able to activate much useful content to effectively participate in the second and third attention computation for better image restoration.

\begin{table*}[!t]\scriptsize
\vspace{-3mm}
\begin{center}
\caption{\textbf{Ablation Study}.
Our full model achieves the best performance in terms of PSNR, SSIM, and MAE.
}
\label{table: Ablation study}
\vspace{-3mm}
\setlength{\tabcolsep}{3.75pt}
\scalebox{1}{
\begin{tabular}{l|l|ccc|cc}
\shline
 ID&Experiment &  PSNR~$\uparrow$ &SSIM~$\uparrow$& MAE~$\downarrow$  & FLOPs (G) & Params (M)\\
\shline
(a)& w/o Nonlinear Activation & 29.359 &0.8765& 0.0311  &46.110& 7.023 \\
(b)& w/o \textit{Value} Nonlinear Trans. Adj. & 29.508 & 0.8786& 0.0306 &41.547&6.388  \\
(c)& w/o Intra Attention Aggregation & 29.627 &0.8805& 0.0303  &39.265&6.071 \\
(d)& w/o Intra Progressive More Heads & 30.101 &0.8875& 0.0283 &46.110& 7.023  \\
(e)& w/o Intra Residual Connections  & 30.189 & 0.8893& 0.0281  &46.110& 7.023 \\
% \textbf{(f)}& w/o Scaling & 0.9207 & 0.1151& 168.91 & 20.48 \\
\shline
\textbf{(f)}& \textbf{Full Model} (\textbf{Ours}) & \textcolor{red}{\textbf{30.391}} &\textcolor{red}{\textbf{0.8903}}& \textcolor{red}{\textbf{0.0276}} &46.110& 7.023 \\
\shline
\end{tabular}}
\end{center}
\vspace{-3mm}
\end{table*}

{\flushleft \bf Effect on Intra Attention Aggregation.}
We aggregate the output features of all the attention in the final.
Tab.~\ref{table: Ablation study} shows that our introduced intra feature aggregation respectively improves PSNR, SSIM, and MAE by $0.764$~dB, 0.98\%, and 0.27\% (Tab.~\ref{table: Ablation study}(c) vs. Tab.~\ref{table: Ablation study}(f)).
The results reveal that aggregating these attention outputs can help the model learn more useful representations with different levels, further improving model performance.

{\flushleft \bf Effect on Intra Progressive More Heads.}
Instead of using the same number of heads in one CSAttn, we suggest using progressive more heads to improve attention ability.
Tab.~\ref{table: Ablation study} shows that our introduced progressive more heads respectively improve PSNR, SSIM, and MAE by $0.290$~dB, 0.28\%, and 0.07\% (Tab.~\ref{table: Ablation study}(d) vs. Tab.~\ref{table: Ablation study}(f)).
The improvement may be because progressive more heads can help the model learn different attention from different levels, which implicitly improves the attention features to further enhance restoration quality.

{\flushleft \bf Effect on Intra Residual Connections.}
We introduce the intra residual connections into continuous attention designs to ease the learning of difference levels.
Tab.~\ref{table: Ablation study} shows that the intra residual connections can improve the PSNR by $0.211$~dB.
By inserting the intra residual connections, our model can provide more useful features into the next attention computation so that it better improves restoration quality.

\begin{figure*}[!t]\scriptsize
\vspace{-5mm}
\begin{center}
\begin{tabular}{cccccccccc}
\includegraphics[width = 0.5\linewidth]{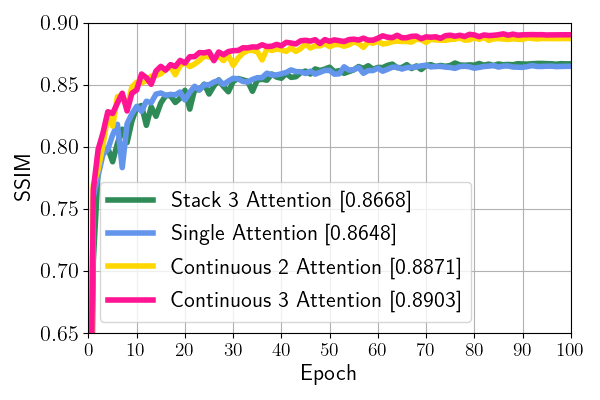}&\hspace{-1.75mm}
\includegraphics[width = 0.5\linewidth]{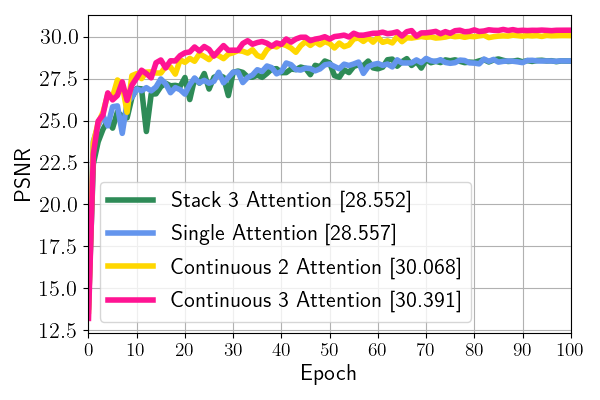}
\end{tabular}
\vspace{-5mm}
\caption{\textbf{Learning Curves on Continuous Attention}.
By plotting the learning curves of single, continuous 2, and continuous 3 attention, we observe that continuous 3 attention outperforms others.
The results reveal that performance can be effectively improved by appropriately utilizing attention within a suitable way.
Note that single attention means the used attention in Restormer~\cite{Zamir2021Restormer}.
}
\label{fig: Effect on Continuous Attention Computation.}
\end{center}
\vspace{-6mm}
\end{figure*}
\begin{figure}[!t]\scriptsize
% \vspace{-1mm}
\begin{center}
\begin{tabular}{cccccccccc}
\includegraphics[width = 0.249\linewidth]{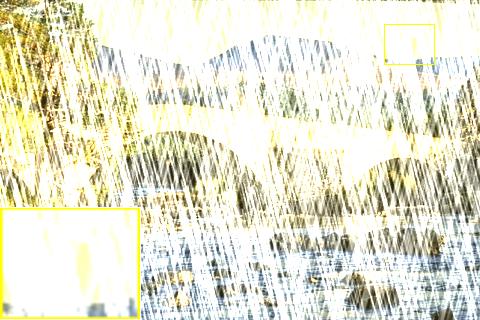}&\hspace{-1.75mm}
\includegraphics[width = 0.249\linewidth]{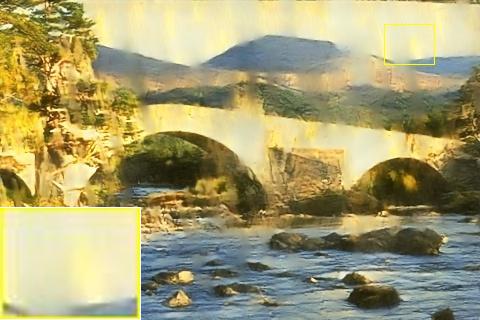}&\hspace{-1.75mm}
\includegraphics[width = 0.249\linewidth]{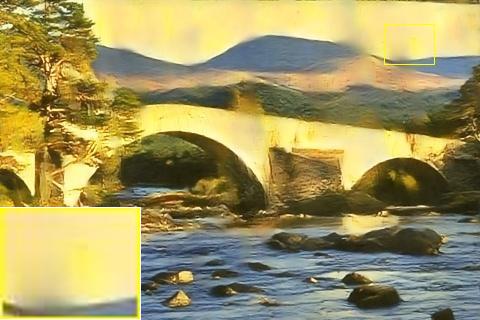}&\hspace{-1.75mm}
\includegraphics[width = 0.249\linewidth]{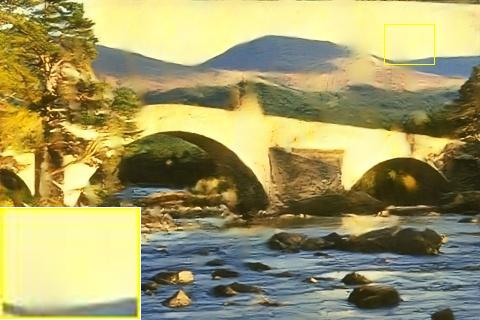}
\\
\includegraphics[width = 0.249\linewidth]{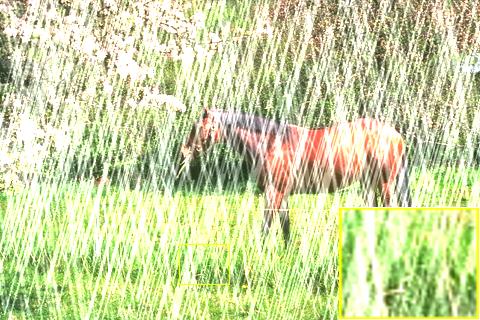}&\hspace{-1.75mm}
\includegraphics[width = 0.249\linewidth]{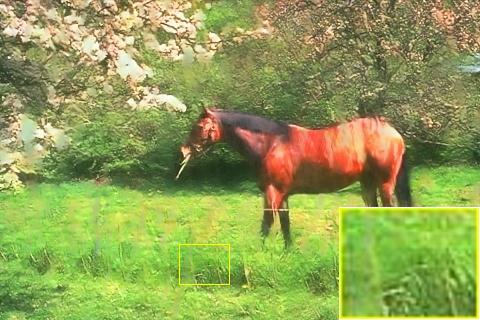}&\hspace{-1.75mm}
\includegraphics[width = 0.249\linewidth]{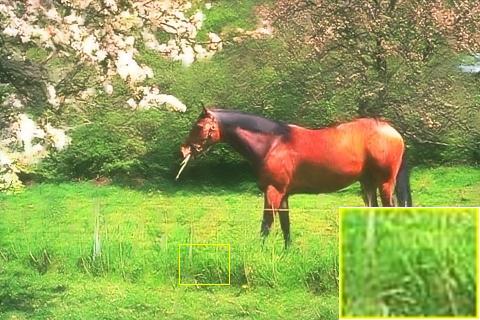}&\hspace{-1.75mm}
\includegraphics[width = 0.249\linewidth]{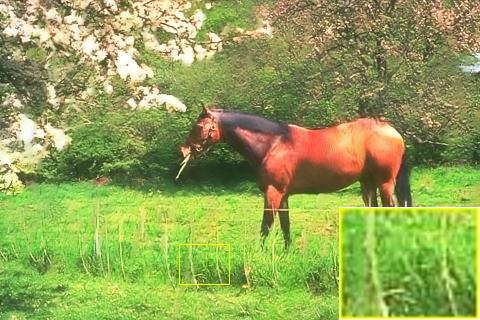}
\\
 Input  &\hspace{-1.75mm}  Single attention  &\hspace{-1.75mm} Continuous 2 attention  &\hspace{-1.75mm}  Continuous 3 attention
\end{tabular}
\vspace{-3mm}
\caption{ \textbf{Visual Effect on Continuous Attention}.
Our continuous 3 attention is able to generate results with fewer artifacts (\textbf{Top}) and finer structures (\textbf{Bottom}), while the single attention is not effective in handling such heavy rain, and the continuous 2 attention hands down artifacts and blur textures.
}
\label{fig: Visual Effect on Continuous Attention Computation.}
\end{center}
\vspace{-6mm}
\end{figure}

\begin{figure}[!t]\scriptsize
% \vspace{-1mm}
\begin{center}
\begin{tabular}{cccccccccc}
\includegraphics[width = 0.249\linewidth]{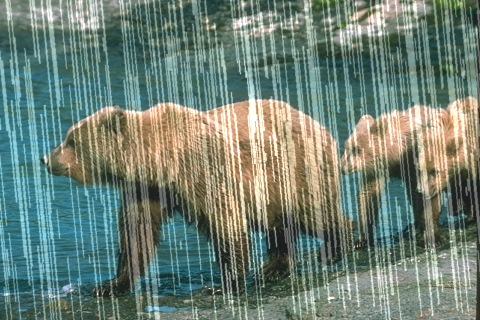}&\hspace{-1.75mm}
\includegraphics[width = 0.249\linewidth]{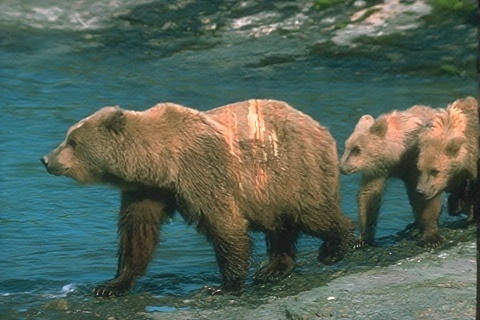}&\hspace{-1.75mm}
\includegraphics[width = 0.249\linewidth]{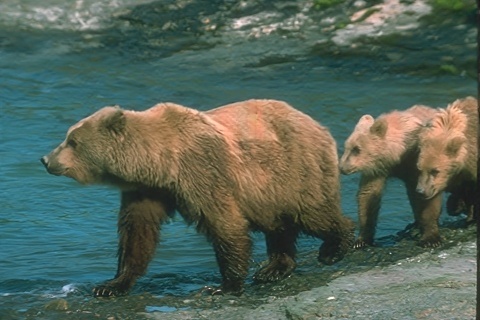}&\hspace{-1.75mm}
\includegraphics[width = 0.249\linewidth]{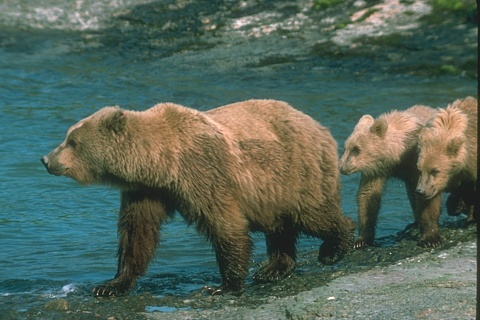}
\\
\includegraphics[width = 0.249\linewidth]{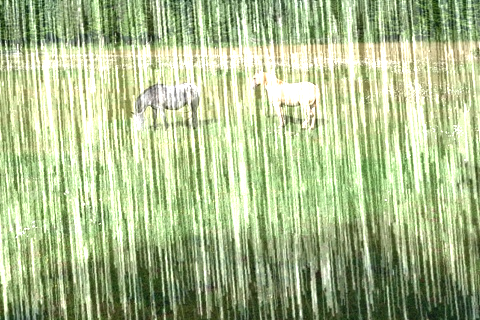}&\hspace{-1.75mm}
\includegraphics[width = 0.249\linewidth]{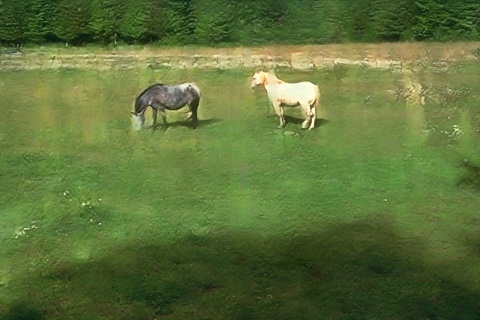}&\hspace{-1.75mm}
\includegraphics[width = 0.249\linewidth]{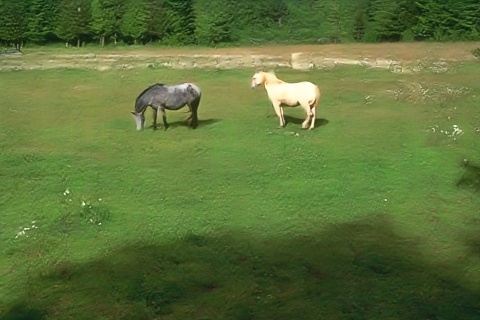}&\hspace{-1.75mm}
\includegraphics[width = 0.249\linewidth]{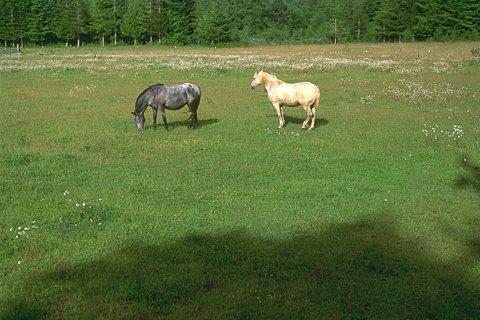}
\\
 Input  &\hspace{-1.75mm}  Stacked 3 Attention   &\hspace{-1.75mm} \textbf{\textit{Ours}}  &\hspace{-1.75mm}  GT
\end{tabular}
\vspace{-3mm}
\caption{ \textbf{Visual Effect on Attention Forms}.
Stacking 3 attention of \cite{Zamir2021Restormer} cannot effectively remove heavy degradation, while our proposed continuous attention form with a series of key designs produces visually pleasing results with fewer artifacts.
%
% Best viewed with zoom-in.
}
\label{fig: Visual Effect on Attention Forms.}
\end{center}
\vspace{-8mm}
\end{figure}

{\flushleft \bf Effect on Continuous Attention.}
We introduce a continuous attention learning mechanism to improve the learning ability of attention.
In this paper, we use continuous $3$ attention to effectively scale up the model capacity.
One may wonder how the different number of continuous attentions affects the restoration quality and what the results of $3$ attention of \cite{Zamir2021Restormer} are compared with the proposed form.
Fig.~\ref{fig: Effect on Continuous Attention Computation.} shows the learning curves of one, continuous two/three attention.
We also compare the model with stacked three attention in \cite{Zamir2021Restormer}.
We note that continuous 3 attention obviously outperforms both 1 attention and continuous 2 attention, which suggests using continuous 3 attention makes attention empower stronger representative learning ability.
Moreover, stacking 3 attention of \cite{Zamir2021Restormer} cannot achieve high-quality restoration results, which further indicates simple cascaded attention is not as good as our designed module.
Fig.~\ref{fig: Visual Effect on Continuous Attention Computation.} presents a visual example of continuous attention with different orders, where our continuous 3 attention is able to generate results with less rainy residuals and finer structures. 
Fig.~\ref{fig: Visual Effect on Attention Forms.} shows that stacked three attention hands down extensive rain traces or artifacts, while our proposed method generates much clearer results.

\begin{figure*}[!t]\scriptsize
\vspace{-3mm}
\begin{center}
\begin{tabular}{cccccccccc}
\includegraphics[width = 0.5\linewidth]{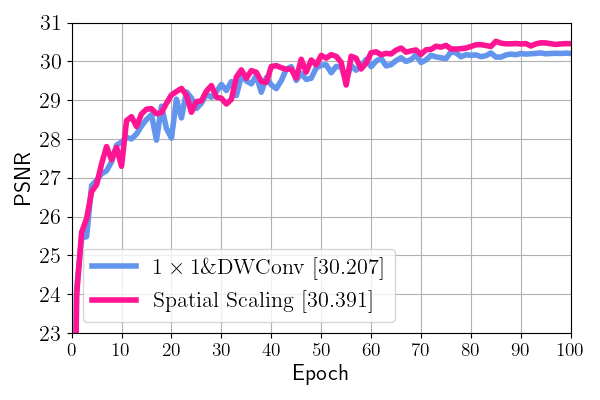}&\hspace{-1.75mm}
\includegraphics[width = 0.5\linewidth]{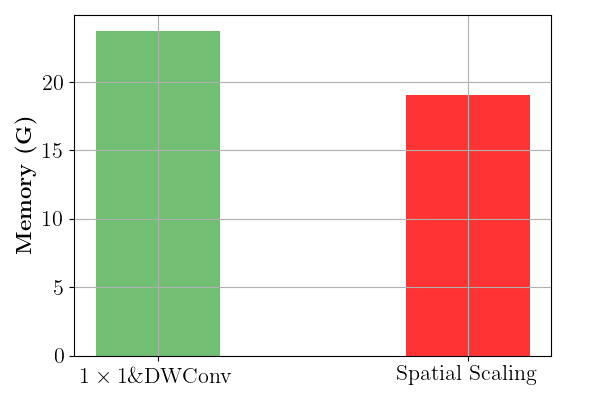}
\\
(a) Learning curves&\hspace{-1.75mm}    (b)  Training Memory
\end{tabular}
\vspace{-3mm}
\caption{\textbf{Learning Curves vs. Training Memory on Spatial Scaling Learning}.
By continuously scaling the spatial sizes into the smaller, the training budget is saved \textbf{(b)} and the results keep competitive with the model without scaling \textbf{(a)}.
}
\label{fig: Effect on Spatial Scaling Learning.}
\end{center}
\vspace{-6mm}
\end{figure*}

\begin{figure}[!t]\scriptsize
% \vspace{-1mm}
\begin{center}
\begin{tabular}{cccccccccc}
\includegraphics[width = 0.249\linewidth]{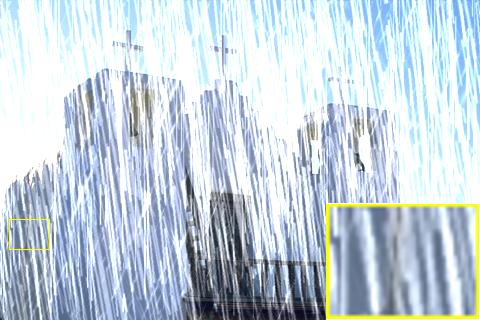}&\hspace{-1.75mm}
\includegraphics[width = 0.249\linewidth]{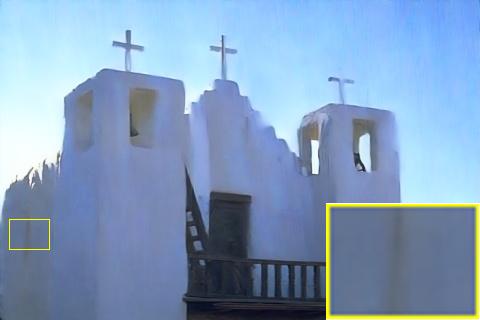}&\hspace{-1.75mm}
\includegraphics[width = 0.249\linewidth]{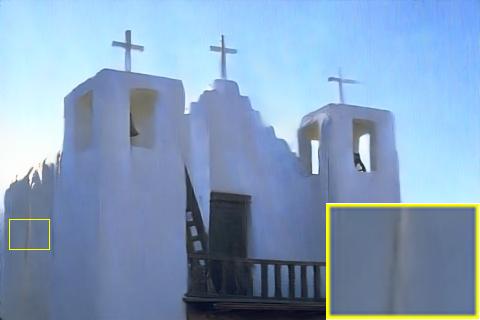}&\hspace{-1.75mm}
\includegraphics[width = 0.249\linewidth]{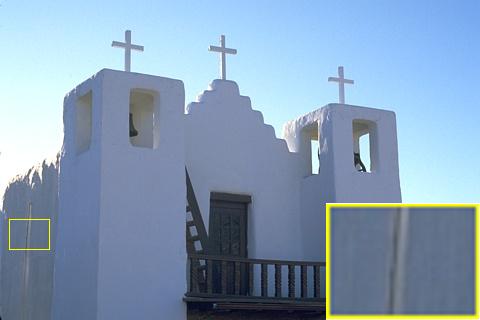}
\\
\includegraphics[width = 0.249\linewidth]{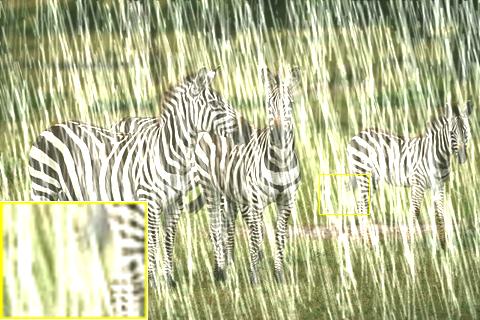}&\hspace{-1.75mm}
\includegraphics[width = 0.249\linewidth]{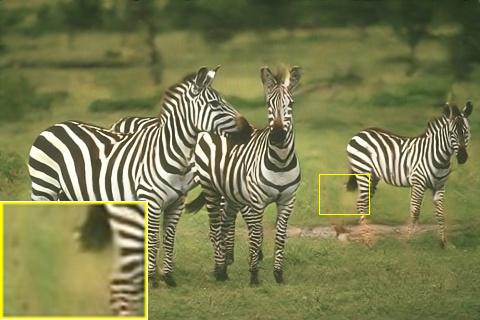}&\hspace{-1.75mm}
\includegraphics[width = 0.249\linewidth]{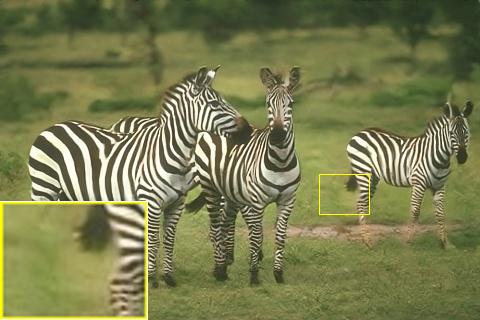}&\hspace{-1.75mm}
\includegraphics[width = 0.249\linewidth]{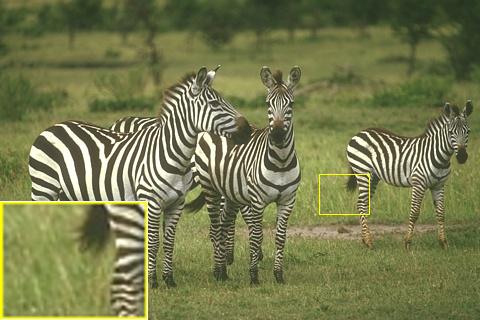}
\\
 Input  &\hspace{-1.75mm} $1 \times 1$\&DWConv.   &\hspace{-1.75mm} Spatial Scaling (\textbf{\textit{Ours}})  &\hspace{-1.75mm}  GT
\end{tabular}
\vspace{-3mm}
\caption{ \textbf{Visual Effect on Spatial Scaling}.
Using spatial scaling is able to generate results with sharper structures and fewer artifacts.
%
% Best viewed with zoom-in.
}
\label{fig: Visual Effect on Spatial Scaling.}
\end{center}
\vspace{-8mm}
\end{figure}

{\flushleft \bf Effect on Spatial Scaling.}
We propose spatial scaling to save the training budget while keeping superior performance compared with the model without using scaling.
Hence, one may be interested in how it affects the results compared model without using spatial scaling.
To answer this question, we compare with the model using $1 \times 1$ convolution and $3 \times 3$ depth-wise convolution which are the same operations with the first attention.
The learning curves are shown in Fig.~\ref{fig: Effect on Spatial Scaling Learning.}(a) and the training memory is illustrated in Fig.~\ref{fig: Effect on Spatial Scaling Learning.}(b).
One can observe that the results of using spatial scaling are better than the model without scaling while consuming less training memory.
Fig.~\ref{fig: Visual Effect on Spatial Scaling.} shows that using spatial scaling is capable of producing results with sharper structures and fewer artifacts.

{\flushleft \bf Effect on Nonlinear Activation Functions.}
Tab.~\ref{table: Effect on nonlinear activation functions} further summarises the results of different nonlinear activation functions.
We note that the GELU activation is the best in terms of PSNR, SSIM, and MAE.
Compared with other activation functions, the PSNR can be improved as large as $0.136$~dB.

\vspace{-3mm}
\subsection{Model Sizes and Computational Complexity}\label{sec:Model Sizes and Computational Complexity}
\vspace{-1mm}
We further report the comparisons of model sizes and computational complexity in Tab.~\ref{table: Model Sizes and Computational Complexity}.
As can be seen, our \textbf{CSAttn} consumes comparable model parameters and FLOPs compared with previous state-of-the-art CNN-based methods and Transformer-based approaches, further demonstrating that the performance improvement is not from the big model sizes.

\begin{table}[!t]\scriptsize
\vspace{-3mm}
\begin{center}
\caption{\textbf{Effect on Nonlinear Activation Functions}.
Compared with ReLU~\cite{glorot2011deep}, LeakyReLU~\cite{maas2013rectifier}, and SiLU~\cite{elfwing2018sigmoid}, GELU~\cite{hendrycks2016gaussian} achieves the best performance in terms of PSNR, SSIM, and MAE.
The PSNR can be improved as large as 0.136dB, compared with ReLU activation.
}
\label{table: Effect on nonlinear activation functions}
\vspace{-3.5mm}
\setlength{\tabcolsep}{5.4pt}
\scalebox{1}{
\begin{tabular}{l|l|ccc|cc}
\shline
 ID&Activation & PSNR~$\uparrow$ & SSIM~$\uparrow$ &MAE~$\downarrow$  & FLOPs (G)~$\downarrow$ & Params (M)~$\downarrow$\\
\shline
(a)&ReLU~\cite{glorot2011deep}       & 30.255& 0.8882 & 0.0280 &46.110& 7.023\\
(b)&LeakyReLU~\cite{maas2013rectifier} &  30.276 &  0.8877 &   0.0279 &46.110& 7.023 \\
(c)&SiLU~\cite{elfwing2018sigmoid}  & 30.286 & 0.8897  & 0.0279 &46.110& 7.023  \\
% \hdashline
\shline
\textbf{(d)}&\textbf{GELU}~\cite{hendrycks2016gaussian} (\textbf{\textit{Default}}) & \textcolor{red}{\textbf{30.391}} &\textcolor{red}{\textbf{0.8903}}& \textcolor{red}{\textbf{0.0276}} &46.110& 7.023 \\
\shline
\end{tabular}}
\end{center}
\vspace{-3mm}
\end{table}
\begin{table}[!t]\scriptsize
% \vspace{-6mm}
\begin{center}
\caption{\textbf{Comparisons on Model Sizes and Computational Complexity}.
Our \textbf{CSAttn} has competitive model sizes and computational complexity compared with state-of-the-art CNN-based methods~\cite{cui2023selective} and Transformer-based approaches~\cite{Zamir2021Restormer,wang2021uformer,PromptRestorer} and Transformer-CNN fusion techniques~\cite{guo2022dehamer}.
}
\label{table: Model Sizes and Computational Complexity}
\vspace{-3mm}
\setlength{\tabcolsep}{6.3pt}
\scalebox{1}{
\begin{tabular}{l|ccccccccccccccccc}
\shline
\multirow{2}{*}{\textbf{Method}} &Restormer & Uformer & DeHamer&SFNet&PromptRestorer&\textbf{CSAttn}\\
&\cite{Zamir2021Restormer}&\cite{wang2021uformer}&\cite{guo2022dehamer}& \cite{cui2023selective}&\cite{PromptRestorer}& \textbf{\textit{Ours}}
\\
\shline
Params (M)~$\downarrow$&  26.1     &20.63&132.5&13.3&24.4 & 21.75\\
FLOPs (G)~$\downarrow$&141.0&41.55& 59.3& 125.4&186.3 & 137.71\\
% \hdashline
\shline
\end{tabular}}
\end{center}
\vspace{-3mm}
\end{table}

\vspace{-3mm}
\section{Concluding Remarks}
\vspace{-1mm}
We have explored the potential of attention for image restoration by proposing a simple and effective Continuous Scaling Attention (\textbf{CSAttn}).
By introducing various design components, we have demonstrated the strong learning ability of attention without using feed-forward networks which are necessary in existing Transformer frameworks.
We have analyzed the effectiveness of each component and revealed its role in our CSAttn.
We have conducted extensive experiments and showed that our CSAttn outperforms state-of-the-art CNN-based and Transformer-based image restoration approaches on various applications, including deraining, desnowing, low-light enhancement, and real dehazing.

\bibliographystyle{splncs04}
\bibliography{example_paper}
\end{document}